\documentclass[10pt,twocolumn,letterpaper]{article}

\usepackage[pagenumbers]{cvpr}      

\usepackage{graphicx}
\usepackage{amsmath}
\usepackage{amssymb}
\usepackage{booktabs}

\usepackage{multirow}
\usepackage{tabularx}
\usepackage{xcolor}
\usepackage{colortbl}
\usepackage[accsupp]{axessibility}
\usepackage[final,tracking=true,kerning=true,spacing=true,factor=1100,stretch=10,shrink=10]{microtype}

\newcommand*{\cm}{\checkmark}
\newcommand*{\z}{\phantom{0}}

\DeclareMathOperator*{\argmax}{arg\,max}

\newcolumntype{Y}{>{\centering\arraybackslash}X}
\newcommand{\isep}{\mathrel{{.}{.}}\nobreak}

\newcommand{\veryshortarrow}[1][3pt]{\mathrel{%
   \hbox{\rule[\dimexpr\fontdimen22\textfont2-.2pt\relax]{#1}{.4pt}}%
   \mkern-4mu\hbox{\usefont{U}{lasy}{m}{n}\symbol{41}}}} 

\usepackage[pagebackref,breaklinks,colorlinks]{hyperref}

\usepackage[capitalize]{cleveref}
\crefname{section}{Sec.}{Secs.}
\Crefname{section}{Section}{Sections}
\Crefname{table}{Table}{Tables}
\crefname{table}{Tab.}{Tabs.}

\begin{document}

\title{MIC: Masked Image Consistency for Context-Enhanced Domain Adaptation}

\author{
   Lukas Hoyer\,\textsuperscript{1} \quad
   Dengxin Dai\,\textsuperscript{2} \quad
   Haoran Wang\,\textsuperscript{2} \quad
   Luc Van Gool\,\textsuperscript{1,3} \\
   \textsuperscript{1}\,ETH Zurich \enskip
   \textsuperscript{2}\,Max Planck Institute for Informatics, Saarland Informatics Campus \enskip
   \textsuperscript{3}\,KU Leuven \\
   {\tt\small \{lhoyer,vangool\}@vision.ee.ethz.ch, \{ddai,hawang\}@mpi-inf.mpg.de}
}
\maketitle

\begin{abstract}
   In unsupervised domain adaptation (UDA), a model trained on source data (e.g. synthetic) is adapted to target data (e.g. real-world) without access to target annotation.
   Most previous UDA methods struggle with classes that have a similar visual appearance on the target domain as no ground truth is available to learn the slight appearance differences. To address this problem, we propose a Masked Image Consistency (MIC) module to enhance UDA by learning spatial context relations of the target domain as additional clues for robust visual recognition.
   MIC enforces the consistency between predictions of masked target images, where random patches are withheld, and pseudo-labels that are generated based on the complete image by an exponential moving average teacher. To minimize the consistency loss, the network has to learn to infer the predictions of the masked regions from their context.
   Due to its simple and universal concept, MIC can be integrated into various UDA methods across different visual recognition tasks such as image classification, semantic segmentation, and object detection. MIC significantly improves the state-of-the-art performance across the different recognition tasks for synthetic-to-real, day-to-nighttime, and clear-to-adverse-weather UDA. For instance, MIC achieves an unprecedented UDA performance of 75.9 mIoU and 92.8\% on GTA$\to$Cityscapes and VisDA-2017, respectively, which corresponds to an improvement of +2.1 and +3.0 percent points over the previous state of the art. The implementation is available at \url{https://github.com/lhoyer/MIC}.
\end{abstract}

\section{Introduction}
\label{sec:intro}

\begin{figure}
    \centering
    \includegraphics[width=.85\linewidth]{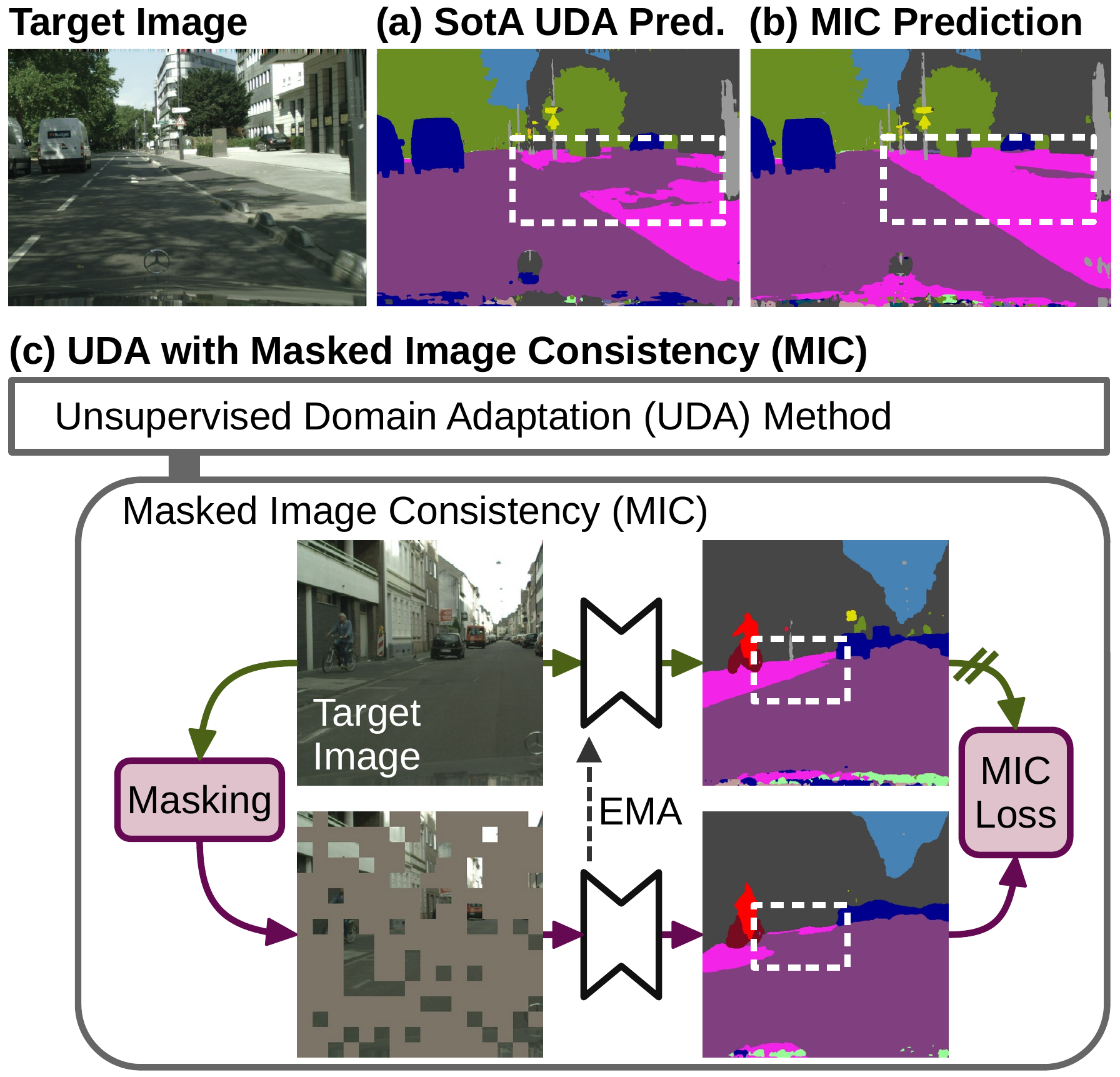}
    \caption{(a) Previous UDA methods such as HRDA~\cite{hoyer2022hrda} struggle with similarly looking classes on the unlabeled target domain. Here, the interior of the \emph{sidewalk} is wrongly segmented as \emph{road}, probably, due to the ambiguous local appearance.
    (b) The proposed Masked Image Consistency (MIC) enhances the learning of context relations to consider additional context clues such as the curb in the foreground. With MIC, the adapted network is able to correctly segment the \emph{sidewalk}.
    (c) MIC can be plugged into most existing UDA methods. It enforces the consistency of the predictions of a masked target image with the pseudo-label of the original image. So, the network is trained to better utilize context clues on the target domain. Further details are shown in Fig.~\ref{fig:method}. 
    }
    \label{fig:overview}
\end{figure}

In order to train state-of-the-art neural networks for visual recognition tasks, large-scale annotated datasets are necessary. However, the collection and annotation process can be very time-consuming and tedious. For instance, the annotation of a single image for semantic segmentation can take more than one hour~\cite{cordts2016cityscapes, sarkadis2021acdc}. Therefore, it would be beneficial to resort to existing or simulated datasets, which are easier to annotate. However, a network trained on such a source dataset usually performs worse when applied to the actual target dataset as neural networks are sensitive to domain gaps. To mitigate this issue, unsupervised domain adaptation (UDA) methods adapt the network to the target domain using unlabeled target images, for instance, with adversarial training~\cite{hoffman2018cycada, tsai2018learning, gong2021dlow, rangwani2022closer} or self-training~\cite{zou2018unsupervised, tranheden2021dacs, wang2021domain, hoyer2021daformer, hoyer2022hrda}.

UDA methods have remarkably progressed in the last few years. However, there is still a noticeable performance gap compared to supervised training. 
A common problem is the confusion of classes with a similar visual appearance on the target domain such as \emph{road}/\emph{sidewalk} or \emph{pedestrian}/\emph{rider} as there is no ground truth supervision available to learn the slight appearance differences. 
For example, the interior of the \emph{sidewalk} in Fig.~\ref{fig:overview} is segmented as \emph{road}, probably, due to a similar local appearance.
To address this problem, we propose to enhance UDA with spatial context relations as additional clues for robust visual recognition. For instance, the curb in the foreground of Fig.~\ref{fig:overview}\,a) could be a crucial context clue to correctly recognize the \emph{sidewalk} despite the ambiguous texture.
Although the used network architectures already have the capability to model context relations, previous UDA methods are still not able to reach the full potential of using context dependencies on the target domain as
the used unsupervised target losses are not powerful enough to enable effective learning of such information.

Therefore, we design a method to explicitly encourage the network to learn comprehensive context relations of the target domain during UDA.
In particular, we propose a novel Masked Image Consistency (MIC) plug-in for UDA (see Fig.~\ref{fig:overview}\,c), which can be applied to various visual recognition tasks. Considering semantic segmentation for illustration, MIC masks out a random selection of target image patches and trains the network to predict the semantic segmentation result of the entire image including the masked-out parts.
In that way, the network has to utilize the context to infer the semantics of the masked regions. As there are no ground truth labels for the target domain, we resort to pseudo-labels, generated by an EMA teacher that uses the original, unmasked target images as input. Therefore, the teacher can utilize both context and local clues to generate robust pseudo-labels. Over the course of the training, different parts of objects are masked out so that the network learns to utilize different context clues, which further increases the robustness. After UDA with MIC, the network is able to better exploit context clues and succeeds in correctly segmenting difficult areas that rely on context clues such as the \emph{sidewalk} in Fig.~\ref{fig:overview}\,b).

\begin{figure}
    \centering
    \includegraphics[width=0.95\linewidth]{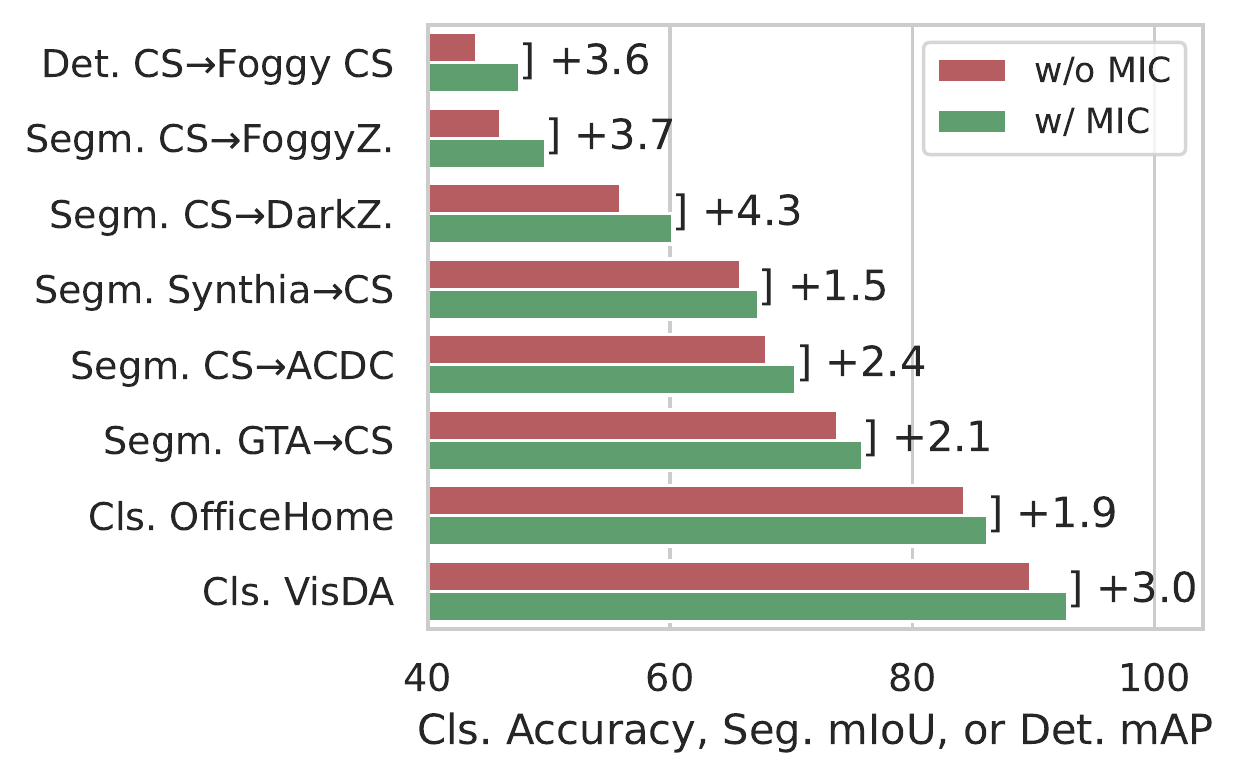}
    \caption{MIC significantly improves state-of-the-art UDA methods across different UDA benchmarks and recognition tasks such as image classification (Cls.), semantic segmentation (Segm.), and object detection (Det.). Detailed results can be found in Sec.~\ref{sec:experiments}.}
    \label{fig:showcase_results}
\end{figure}

To the best of our knowledge, MIC is the first UDA approach to exploit masked images to facilitate learning context relations on the target domain. Due to its universality and simplicity, MIC can be straightforwardly integrated into various UDA methods across different visual recognition tasks, making it highly valuable in practice. MIC achieves significant and consistent performance improvements for different UDA methods (including adversarial training, entropy-minimization, and self-training) on multiple visual recognition tasks (image classification, semantic segmentation, and object detection) with different domain gaps (synthetic-to-real, clear-to-adverse-weather, and day-to-night) and different network architectures (CNNs and Transformer). It sets a new state-of-the-art performance on all tested benchmarks with significant improvements over previous methods as shown in Fig.~\ref{fig:showcase_results}. For instance, MIC respectively improves the state-of-the-art performance by +2.1, +4.3, and +3.0 percent points on GTA$\to$Cityscapes(CS), CS$\to$DarkZurich, and VisDA-2017 and achieves an unprecedented UDA performance of 75.9 mIoU, 60.2 mIoU, and 92.8\%, respectively.

\section{Related Work}
\label{sec:related}

\subsection{Unsupervised Domain Adaptation (UDA)}

In UDA, a model trained on a labeled source domain is adapted to an unlabeled target domain. Due to the ubiquity of domain gaps, UDA methods were designed for all major computer vision problems including
image classification~\cite{long2015learning,ganin2016domain,long2018conditional,pan2019transferrable}, semantic segmentation~\cite{hoffman2016fcns,tsai2018learning,zhang2021prototypical,hoyer2021daformer}, and object detection~\cite{chen2018domain, chen2021scale, li2022cross, li2022sigma}.
The majority of the approaches rely on discrepancy minimization, adversarial training, or self-training. 
The first group minimizes the discrepancy between domains using a statistical distance function such as maximum mean discrepancy~\cite{gretton2006kernel,long2015learning,long2017deep}, correlation alignment~\cite{sun2016return,sun2016deep}, or entropy minimization~\cite{grandvalet2004semi,long2016unsupervised,vu2019advent}.
In adversarial training, a learned domain discriminator provides supervision in a GAN framework~\cite{goodfellow2014generative} to encourage domain-invariant inputs~\cite{hoffman2018cycada,gong2019dlow}, features~\cite{ganin2016domain,hoffman2016fcns,long2018conditional,tsai2018learning} or outputs~\cite{saito2018maximum,tsai2018learning,vu2019advent,luo2021category}.
In self-training, pseudo-labels~\cite{lee2013pseudo} are generated for the target domain based on predictions obtained using confidence thresholds~\cite{zhang2018collaborative,zou2018unsupervised,mei2020instance} or pseudo-label prototypes~\cite{pan2019transferrable,zhang2019category,zhang2021prototypical}. To increase the robustness of the self-training, consistency regularization~\cite{sajjadi2016regularization, tarvainen2017mean, sohn2020fixmatch} is often applied to ensure consistency over different data augmentations~\cite{french2017self,choi2019self, melaskyriazi2021pixmatch, araslanov2021self}, different crops~\cite{lai2021semi,hoyer2022hrda}, multiple models~\cite{zhou2020uncertainty,zheng2021rectifying,zhang2021multiple}, or domain-mixup~\cite{tranheden2021dacs, zhou2021context,hoyer2021improving,hoyer2021daformer,hoyer2022hrda}. 
Further UDA strategies utilize pretext tasks~\cite{vu2019dada,chen2019learning,wang2021domain,hoyer2021improving}, follow an adaptation curriculum~\cite{dai2018dark,zhang2019curriculum,dai2020curriculum}, exploit the increased domain-robustness of Transformers~\cite{hoyer2021daformer,xu2021cdtrans,sun2022safe,hoyer2022hrda}, align the domains with contrastive learning~\cite{xie2022sepico,huang2022category}, use graph matching~\cite{cai2019exploring, li2022scan,li2022sigma}, or adapt multi-resolution inputs~\cite{hoyer2022hrda}.

To facilitate learning domain-robust context dependencies, several UDA methods propose network components that can capture context such as spatial attention pyramids~\cite{li2020spatial}, cross-domain attention~\cite{yang2021context}, or context-aware feature fusion~\cite{hoyer2021daformer}. 
While these network modules provide the ability to capture context, the unsupervised loss on the target domain does not provide sufficient supervision to learn all relevant target context relations. 
To improve context learning, CrDA~\cite{huang2020contextual} aligns local context relations with adversarial training and HRDA~\cite{hoyer2022hrda} uses multi-crop consistency training. However, these mechanisms are not able to capture all relevant context clues as can be seen for HRDA in Fig.~\ref{fig:overview}\,a).
Due to the random patch masking, MIC is able to learn a larger set of different context clues for robust recognition.

\subsection{Masked Image Modeling}

Predicting withheld tokens of a masked input sequence was shown to be a powerful self-supervised pretraining task in natural language processing~\cite{devlin2018bert, brown2020language}. 
Recently, this concept was successfully transferred to self-supervised pretraining in computer vision, where it is known as masked image modeling. Given a partly masked image, the network is trained to reconstruct properties of the masked areas such as VAE features~\cite{bao2021beit,dong2021peco,li2022mc}, HOG features~\cite{wei2022masked}, or color information~\cite{he2022masked,xie2022simmim}. To sample the mask, block-wise masking~\cite{bao2021beit}, random patch masking~\cite{he2022masked,xie2022simmim}, and attention-guided masking~\cite{li2021mst,kakogeorgiou2022hide} have been explored.

Similarly, our method also uses masked images. However, we pursue a different purpose than previous works. Instead of aiming to learn self-supervised representations, MIC utilizes masked images in a novel way to learn context relations for domain adaptation. Due to this conceptual difference, we do not have to rely on pretext restoration targets such as VAE features but can perform the reconstruction in the actual prediction space of the relevant computer vision task such as semantic segmentation. To the best of our knowledge, MIC is the first method to exploit masked images to enhance context learning for UDA.
Particularly, we show that naive masked image modeling on ImageNet does not improve the target domain performance (see Sec.~\ref{sec:exp_classification}).

\section{Methods}
\label{sec:methods}

\begin{figure}
    \centering
    \includegraphics[width=\linewidth]{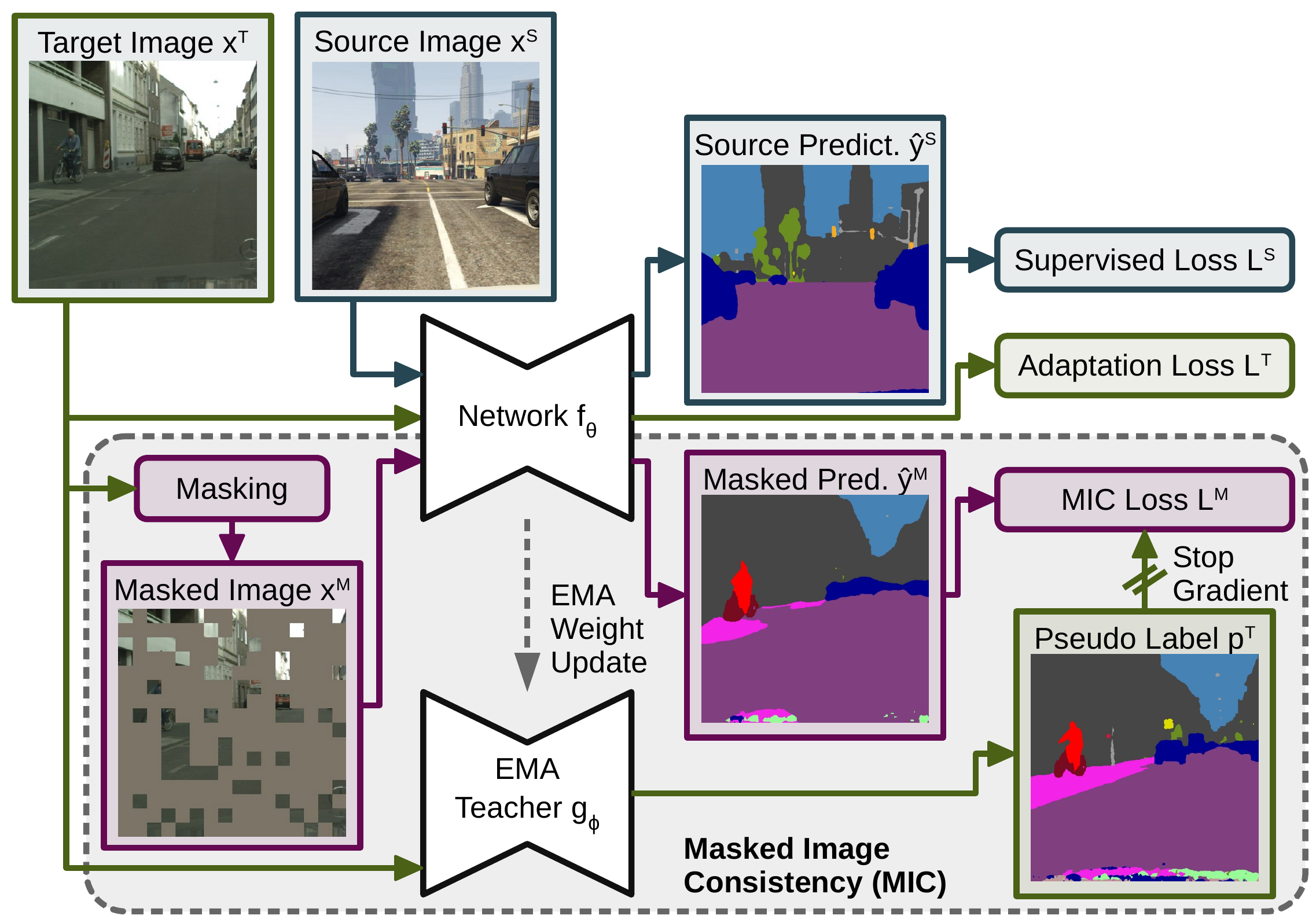}
    \caption{UDA with the proposed Masked Image Consistency (MIC). In UDA, a network is typically trained with a supervised loss on the source domain (blue) and an unsupervised adaptation loss on the target domain (green). MIC enforces the consistency between predictions of masked target images (purple) and pseudo-labels that are generated based on the complete image by an exponential moving average (EMA) teacher. To minimize the MIC loss, the network has to learn to infer the predictions of the masked regions from their context. 
    }
    \label{fig:method}
\end{figure}

\subsection{Unsupervised Domain Adaptation (UDA)}

A neural network $f_\theta$ can be trained on the source domain using images $\mathcal{X}^S = \{x_{k}^{S}\}_{k=1}^{N_S}$ and their labels $\mathcal{Y}^S = \{y_k^{S}\}_{k=1}^{N_S}$ with a supervised source loss $\mathcal{L}^S$. The specific source loss depends on the computer vision task. For image classification and semantic segmentation, the (pixel-wise) cross-entropy is typically used
\begin{align}
    \mathcal{L}^\mathit{S,cls/seg}_k &= \mathcal{H}(f_\theta(x^S_k), y^S_k)\,,\\
    \mathcal{H}(\hat{y}, y) &= - \sum_{i=1}^{H} \sum_{j=1}^{W} \sum_{c=1}^C y_{ijc} \log \hat{y}_{ijc}\,,
\end{align}
where $H{=}W{=}1$ in case of classification. For object detection, a box regression and a box classification loss are commonly utilized~\cite{ren2015faster}.

However, a model trained on the source domain usually experiences a performance drop when applied to another domain. Therefore, unsupervised domain adaptation (UDA) methods use unlabeled images from the target domain $\mathcal{X}^T= \{x_{k}^{T}\}_{k=1}^{N_T}$ to adapt the network.
For that purpose, an additional unsupervised loss for the target domain $\mathcal{L}^T$ is added to the optimization problem with a weight $\lambda^T$
\begin{equation}
    \min_{\theta} \frac{1}{N_S} \sum_{k=1}^{N_S} \mathcal{L}^S_k + \frac{1}{N_T} \sum_{k=1}^{N_T} \lambda^T\mathcal{L}_k^T\;.
\end{equation}

The target loss $\mathcal{L}^T$ is defined according to the UDA strategy such as adversarial training~\cite{ganin2016domain,tsai2018learning,tsai2019domain,wang2020classes, chen2021scale, rangwani2022closer} or self-training~\cite{zou2018unsupervised,zhang2019category,mei2020instance,tranheden2021dacs,zhang2021prototypical,hoyer2021daformer}. 

 \subsection{Masked Image Consistency (MIC)}

To recognize an object (or stuff region), a model can utilize clues from different parts of the image. This can be local information, which originates from the same image patch as the corresponding cell in the feature map, or context information, which comes from surrounding image patches that can belong to different parts of the object or its environment~\cite{hoyer2019grid}. Many network architectures~\cite{he2016deep,dosovitskiy2020image} have the capability to integrate both local and context information in their features.
While the learning of context clues can be guided by ground truth in supervised learning, there is no ground truth supervision available for the target domain in UDA.
Current unsupervised losses are not powerful enough to enable effective learning of context clues as empirically observed such as in Fig.~\ref{fig:overview}\,a). Therefore, we propose to specifically encourage the learning of context relations on the target domain to provide additional clues for robust recognition of classes with similar local appearances.

In order to facilitate the learning of context relations on the target domain, we introduce a Masked Image Consistency (MIC) module, which can be easily plugged into various existing UDA methods. The domain adaptation process with MIC is illustrated in Fig.~\ref{fig:method} and explained below.

MIC withholds local information by randomly masking out patches of the target image. For that purpose, a patch mask $\mathcal{M}$ is randomly sampled from a uniform distribution
\begin{equation}
    \mathcal{M}_{\substack{mb+1:(m+1)b,\\nb+1:(n+1)b}} = [v > r] \quad \text{with} \; v \sim \mathcal{U}(0, 1)\;,
\end{equation}
where $[\cdot]$ denotes the Iverson bracket, $b$ the patch size, $r$ the mask ratio, and $m\in[0 \isep W/b-1]$, $n\in[0 \isep W/b-1]$ the patch indices. The masked target image $x^M$ (see Fig.~\ref{fig:method}) is obtained by element-wise multiplication of mask and image
\begin{equation}
    x^{M} = \mathcal{M} \odot x^{T}\,.
\end{equation}
The masked target prediction $\hat{y}^{M}$ can only use the limited information of the unmasked image regions
\begin{equation}
    \hat{y}^{M} = f_\theta(x^M)\,,
\end{equation}
making the prediction more difficult. This is also reflected in Fig.~\ref{fig:method}, where the prediction misses a part of the sidewalk. 
In order to train the network to use the remaining context clues to still reconstruct the correct label without access to the entire image, the MIC loss $\mathcal{L}^M$ is introduced
\begin{equation}
    \mathcal{L}^M =  q^T \mathcal{H}(\hat{y}^M, p^T)\,,
\end{equation}
where $p^T$ denotes a pseudo-label and $q^T$ its quality weight. 
MIC uses pseudo-labels as there is no ground truth available for the target domain. The pseudo-label is the prediction of a teacher network $g_\phi$ of the complete target image $x^T$. For image classification and semantic segmentation,
\begin{equation}
    p^\mathit{T,cls/seg}_{ij} = [c = \argmax_{c'} g_\phi(x^T)_{ijc'}]\,.
    \label{eq:pseudo_label}
\end{equation}
For object detection pseudo-labels, box predictions from $g_\phi(x^T)$ are filtered with a confidence threshold $\delta$ and non-maximum suppression~\cite{ren2015faster}.

The teacher network $g_\phi$ is implemented as an EMA teacher~\cite{tarvainen2017mean}. Its weights are the exponential moving average of the weights of $f_\theta$ with smoothing factor $\alpha$
\begin{equation}
    \phi_{t+1} \leftarrow \alpha \phi_t + (1 - \alpha) \theta_t\,,
\end{equation}
where $t$ denotes a training step. The EMA teacher realizes a temporal ensemble of previous student models $f_\theta$~\cite{tarvainen2017mean
}, which increases the robustness and temporal stability of pseudo-labels. It is a common strategy used in semi-supervised learning~\cite{tarvainen2017mean,french2019consistency,hoyer2021three} and UDA~\cite{araslanov2021self,tranheden2021dacs,hoyer2021daformer,hoyer2022hrda}.
As the teacher is updated based on the student $f_\theta$, it will gradually obtain the enhanced context learning capability from $f_\theta$. In contrast to the student $f_\theta$, the teacher $g_\phi$ has privileged access to the original image $x^T$ (see Eq.~\ref{eq:pseudo_label}), 
so that it can use both the context and the intact local appearance information to generate pseudo labels of higher quality.

As the pseudo-labels are potentially wrong (especially at the beginning of the training), the loss is weighted by the quality estimate $q^T$. For image classification, we use the maximum softmax probability as certainty estimate~\cite{zhang2018collaborative}
\begin{equation}
    q^\mathit{T,cls} = \max_{c'} g_\phi(x^T)_{c'}\,.
    \label{eq:q_classification}
\end{equation}
For semantic segmentation, we follow \cite{tranheden2021dacs,hoyer2021daformer,hoyer2022hrda} and utilize the ratio of pixels exceeding a threshold $\tau$ of the maximum softmax probability
\begin{equation}
    q^\mathit{T,seg} = \frac{\sum_{i=1}^{H}\sum_{j=1}^{W} [\max_{c'} g_\phi(x^T)_{(ijc')} > \tau]}{H \cdot W}\,.
\end{equation}
And for object detection, we apply the quality estimate from Eq.~\ref{eq:q_classification} to each bounding box in the classification branches.

The MIC consistency training can be easily integrated into the UDA optimization problem \begin{equation}
    \min_{\theta} \frac{1}{N_S} \sum_{k=1}^{N_S} \mathcal{L}^S_k + \frac{1}{N_T} \sum_{k=1}^{N_T} (\lambda^T \mathcal{L}_k^T + \lambda^M \mathcal{L}_k^M)\;.
\end{equation}

\section{Experiments}
\label{sec:experiments}

\subsection{Implementation Details}
\label{sec:implementation}

\noindent\textbf{Semantic Segmentation:}
We study synthetic-to-real, clear-to-adverse-weather, and day-to-nighttime adaptation of street scenes. 
As synthetic datasets, we use GTA~\cite{richter2016playing} containing 24,966 images and Synthia~\cite{ros2016synthia} with 9,400 images. As real-world datasets, we use Cityscapes (CS)~\cite{cordts2016cityscapes} consisting of 2,975 training and 500 validation images for clear weather, DarkZurich~\cite{sakaridis2020map} with 2,416 training
and 151 test images for nighttime, and ACDC~\cite{sarkadis2021acdc} containing 1,600 training, 406 validation, and 2,000 test images for adverse weather (fog, night, rain, and snow). The training resolution follows the used UDA methods (e.g. half resolution for DAFormer~\cite{hoyer2021daformer} or full resolution for HRDA~\cite{hoyer2022hrda}).

We evaluate MIC based on a DAFormer network~\cite{hoyer2021daformer} with a MiT-B5 encoder~\cite{xie2021segformer}, and a DeepLabV2~\cite{chen2017deeplab} with a ResNet-101~\cite{he2016deep} backbone. All backbones are initialized with ImageNet pretraining.
In the default UDA setting, we follow the HRDA~\cite{hoyer2022hrda} multi-resolution self-training strategy and training parameters, i.e. AdamW~\cite{loshchilov2018decoupled} with a learning rate of $6 {\times} 10^{-5}$ for the encoder and $6 {\times} 10^{-4}$ for the decoder, 40k training iterations, a batch size of 2, linear learning rate warmup, a loss weight $\lambda^T_\mathit{st}{=}1$, an EMA factor $\alpha{=}0.999$, DACS~\cite{tranheden2021dacs} data augmentation, Rare Class Sampling~\cite{hoyer2021daformer}, and ImageNet Feature Distance~\cite{hoyer2021daformer}. 
For adversarial training and entropy minimization, SGD with a learning rate of $0.0025$ and $\lambda^T_\mathit{adv}{=}\lambda^T_\mathit{ent}{=}0.001$ is used.

\noindent\textbf{Image Classification:}
We evaluate MIC on the VisDA-2017 dataset~\cite{peng2017visda}, which consists of 280,000 synthetic and real images of 12 classes, as well as the Office-Home dataset~\cite{venkateswara2017deep}, which contains 15,500 images from 65 classes for the domains art (A), clipart (C), product (P) and real-world (R).
We conduct the experiments with ResNet-101~\cite{he2016deep} and ViT-B/16~\cite{dosovitskiy2020image}. For UDA training, we follow SDAT~\cite{rangwani2022closer}, which utilizes CDAN~\cite{long2018conditional} with MCC~\cite{jin2020minimum} and a smoothness enhancing loss~\cite{rangwani2022closer}. We use the same training parameters, i.e. SGD with a learning rate of 0.002, a batch size of 32, and a smoothness parameter of 0.02.

\noindent\textbf{Object Detection:}
For object detection UDA, we evaluate MIC on CS~\cite{cordts2016cityscapes} to Foggy CS~\cite{sakaridis2018semantic}.
The experiments are performed based on Faster R-CNN \cite{ren2015faster} with ResNet-50 \cite{he2016deep} and FPN \cite{lin2017feature}. For UDA, we adopt SADA \cite{chen2021scale}, which utilizes adversarial training 
on image and instance level. The same parameters as in~\cite{chen2021scale} are used, i.e. 0.0025 initial learning rate, 60k training iterations, $\lambda^T_\mathit{adv}{=}0.1$, and a batch size of 2. Following previous works \cite{saito2019strong,chen2021scale}, we report the results in mean Average Precision (mAP) with a 0.5 IoU threshold.

\noindent\textbf{MIC Parameters:}
MIC uses a patch size $b{=}64$, a mask ratio $r{=}0.7$, a loss weight $\lambda^M{=}1$, an EMA weight $\alpha{=}0.999$ following~\cite{hoyer2021daformer,hoyer2022hrda}, and color augmentation (brightness, contrast, saturation, hue, and blur) following the parameters of~\cite{tranheden2021dacs,hoyer2021daformer,hoyer2022hrda}.
We set the pseudo-label box threshold $\delta{=}0.8$ following~\cite{deng2021unbiased, li2022cross} and the quality threshold $\tau{=}0.968$ following~\cite{tranheden2021dacs,hoyer2021daformer,hoyer2022hrda}. 
If a UDA method trains with half resolution~\cite{tsai2018learning,vu2019advent,chen2021scale,tranheden2021dacs,hoyer2021daformer}, the patch size is divided by 2.
For image classification and object detection, we use $\alpha{=}0.9$.
For object detection, we reduce the mask ratio $r{=}0.5$ as
the objects of interest are more sparse and a high $r$ increases the risk that they are completely masked out.
For target domains with nighttime images (DarkZurich and ACDC), we forgo color augmentation as it can corrupt the content of dark nighttime images due to the locally already low brightness and contrast.
The experiments are conducted on an RTX 2080 Ti or a Titan RTX depending on the required memory.

\subsection{MIC for Semantic Segmentation}

\begin{table}
\centering
\caption{Segmentation performance (mIoU in \%) of MIC with different UDA methods on GTA$\to$CS.}
\label{tab:other_uda}
\setlength{\tabcolsep}{5pt}
\scriptsize
\begin{tabular}{llccc}
\hline
                           Network &                          UDA Method & w/o MIC & w/ MIC & Diff. \\
\hline\hline
DeepLabV2 \cite{chen2017deeplab} & Adversarial \cite{tsai2018learning} &    44.2 &   48.2 &        +4.0 \\
DeepLabV2 \cite{chen2017deeplab} &    Entropy Min. \cite{vu2019advent} &    44.3 &   49.0 &        +4.7 \\
DeepLabV2 \cite{chen2017deeplab} &       DACS \cite{tranheden2021dacs} &    53.9 &   56.0 &        +2.1 \\
DeepLabV2 \cite{chen2017deeplab} &   DAFormer \cite{hoyer2021daformer} &    56.0 &   59.4 &        +3.4 \\
DeepLabV2 \cite{chen2017deeplab} &           HRDA \cite{hoyer2022hrda} &    63.0 &   64.2 &        +1.2 \\
\hline
DAFormer \cite{hoyer2021daformer} &   DAFormer \cite{hoyer2021daformer} &    68.3 &   70.6 &        +2.3 \\
DAFormer \cite{hoyer2021daformer} &           HRDA \cite{hoyer2022hrda} &    73.8 &   75.9 &        +2.1 \\
\hline
\end{tabular}
\end{table}

First, we combine MIC with different UDA methods and network architectures for semantic segmentation on GTA$\to$CS. Tab.~\ref{tab:other_uda} shows that MIC achieves consistent and significant improvements across various UDA methods with different network architectures, ranging from +1.2 up to +4.7 mIoU. Specifically, MIC does not only benefit powerful Transformers such as DAFormer~\cite{hoyer2021daformer} but also CNNs such as DeepLabV2~\cite{chen2017deeplab}. Across UDA methods, the performance improvement decreases with a higher UDA performance as expected due to performance saturation.

\begin{table*}
\centering
\caption{Semantic segmentation performance (IoU in \%) on four different UDA benchmarks.}
\label{tab:sota_segmentation}
\setlength{\tabcolsep}{3pt}
\scriptsize
\begin{tabular}{l|ccccccccccccccccccc|c}
\hline
Method & Road & S.walk & Build. & Wall & Fence & Pole & Tr.Light & Sign & Veget. & Terrain & Sky & Person & Rider & Car & Truck & Bus & Train & M.bike & Bike & mIoU\\
\toprule
\multicolumn{21}{c}{\textbf{Synthetic-to-Real: GTA$\to$Cityscapes (Val.)}} \\
\hline
ADVENT~\cite{vu2019advent} & 89.4 & 33.1 & 81.0 & 26.6 & 26.8 & 27.2 & 33.5 & 24.7 & 83.9 & 36.7 & 78.8 & 58.7 & 30.5 & 84.8 & 38.5 & 44.5 & 1.7 & 31.6 & 32.4 & 45.5\\
DACS~\cite{tranheden2021dacs} & 89.9 & 39.7 & 87.9 & 30.7 & 39.5 & 38.5 & 46.4 & 52.8 & 88.0 & 44.0 & 88.8 & 67.2 & 35.8 & 84.5 & 45.7 & 50.2 & 0.0 & 27.3 & 34.0 & 52.1\\
ProDA~\cite{zhang2021prototypical} & 87.8 & 56.0 & 79.7 & 46.3 & 44.8 & 45.6 & 53.5 & 53.5 & 88.6 & 45.2 & 82.1 & 70.7 & 39.2 & 88.8 & 45.5 & 59.4 & 1.0 & 48.9 & 56.4 & 57.5\\
DAFormer~\cite{hoyer2021daformer} & 95.7 & 70.2 & 89.4 & 53.5 & 48.1 & 49.6 & 55.8 & 59.4 & 89.9 & 47.9 & 92.5 & 72.2 & 44.7 & 92.3 & 74.5 & 78.2 & 65.1 & 55.9 & 61.8 & 68.3\\
HRDA~\cite{hoyer2022hrda} & \underline{96.4} & \underline{74.4} & \underline{91.0} & \textbf{61.6} & \underline{51.5} & \underline{57.1} & \underline{63.9} & \underline{69.3} & \underline{91.3} & \underline{48.4} & \underline{94.2} & \underline{79.0} & \underline{52.9} & \underline{93.9} & \underline{84.1} & \underline{85.7} & \underline{75.9} & \underline{63.9} & \underline{67.5} & \underline{73.8}\\
MIC~(HRDA) & \textbf{97.4} & \textbf{80.1} & \textbf{91.7} & \underline{61.2} & \textbf{56.9} & \textbf{59.7} & \textbf{66.0} & \textbf{71.3} & \textbf{91.7} & \textbf{51.4} & \textbf{94.3} & \textbf{79.8} & \textbf{56.1} & \textbf{94.6} & \textbf{85.4} & \textbf{90.3} & \textbf{80.4} & \textbf{64.5} & \textbf{68.5} & \textbf{75.9}\\
\toprule
\multicolumn{21}{c}{\textbf{Synthetic-to-Real: Synthia$\to$Cityscapes (Val.)}} \\
\hline

ADVENT~\cite{vu2019advent} & 85.6 & 42.2 & 79.7 & 8.7 & 0.4 & 25.9 & 5.4 & 8.1 & 80.4 & -- & 84.1 & 57.9 & 23.8 & 73.3 & -- & 36.4 & -- & 14.2 & 33.0 & 41.2\\
DACS~\cite{tranheden2021dacs} & 80.6 & 25.1 & 81.9 & 21.5 & 2.9 & 37.2 & 22.7 & 24.0 & 83.7 & -- & 90.8 & 67.6 & 38.3 & 82.9 & -- & 38.9 & -- & 28.5 & 47.6 & 48.3\\
ProDA~\cite{zhang2021prototypical} & \textbf{87.8} & 45.7 & 84.6 & 37.1 & 0.6 & 44.0 & 54.6 & 37.0 & \textbf{88.1} & -- & 84.4 & 74.2 & 24.3 & 88.2 & -- & 51.1 & -- & 40.5 & 45.6 & 55.5\\
DAFormer~\cite{hoyer2021daformer} & 84.5 & 40.7 & 88.4 & 41.5 & \underline{6.5} & 50.0 & 55.0 & 54.6 & 86.0 & -- & 89.8 & 73.2 & 48.2 & 87.2 & -- & 53.2 & -- & 53.9 & 61.7 & 60.9\\
HRDA~\cite{hoyer2022hrda} & 85.2 & \underline{47.7} & \underline{88.8} & \textbf{49.5} & 4.8 & \underline{57.2} & \underline{65.7} & \underline{60.9} & 85.3 & -- & \underline{92.9} & \underline{79.4} & \underline{52.8} & \underline{89.0} & -- & \textbf{64.7} & -- & \underline{63.9} & \textbf{64.9} & \underline{65.8}\\
MIC~(HRDA) & \underline{86.6} & \textbf{50.5} & \textbf{89.3} & \underline{47.9} & \textbf{7.8} & \textbf{59.4} & \textbf{66.7} & \textbf{63.4} & \underline{87.1} & -- & \textbf{94.6} & \textbf{81.0} & \textbf{58.9} & \textbf{90.1} & -- & \underline{61.9} & -- & \textbf{67.1} & \underline{64.3} & \textbf{67.3}\\

\toprule
\multicolumn{21}{c}{\textbf{Day-to-Nighttime: Cityscapes$\to$DarkZurich (Test)}} \\
\hline

ADVENT~\cite{vu2019advent} & 85.8 & 37.9 & 55.5 & 27.7 & 14.5 & 23.1 & 14.0 & 21.1 & 32.1 & 8.7 & 2.0 & 39.9 & 16.6 & 64.0 & 13.8 & 0.0 & 58.8 & 28.5 & 20.7 & 29.7\\
MGCDA$^\dagger$~\cite{sakaridis2020map} & 80.3 & 49.3 & 66.2 & 7.8 & 11.0 & 41.4 & 38.9 & 39.0 & \underline{64.1} & 18.0 & 55.8 & 52.1 & 53.5 & 74.7 & \underline{66.0} & 0.0 & 37.5 & 29.1 & 22.7 & 42.5\\
DANNet$^\dagger$~\cite{wu2021dannet} & 90.0 & 54.0 & \underline{74.8} & \underline{41.0} & \underline{21.1} & 25.0 & 26.8 & 30.2 & \textbf{72.0} & 26.2 & \textbf{84.0} & 47.0 & 33.9 & 68.2 & 19.0 & 0.3 & 66.4 & 38.3 & 23.6 & 44.3\\
DAFormer~\cite{hoyer2021daformer} & \underline{93.5} & \underline{65.5} & 73.3 & 39.4 & 19.2 & 53.3 & \underline{44.1} & \underline{44.0} & 59.5 & \underline{34.5} & 66.6 & 53.4 & 52.7 & \underline{82.1} & 52.7 & 9.5 & 89.3 & 50.5 & 38.5 & 53.8\\
HRDA~\cite{hoyer2022hrda} & 90.4 & 56.3 & 72.0 & 39.5 & 19.5 & \underline{57.8} & \textbf{52.7} & 43.1 & 59.3 & 29.1 & \underline{70.5} & \underline{60.0} & \underline{58.6} & \textbf{84.0} & \textbf{75.5} & \underline{11.2} & \underline{90.5} & \underline{51.6} & \underline{40.9} & \underline{55.9}\\
MIC~(HRDA) & \textbf{94.8} & \textbf{75.0} & \textbf{84.0} & \textbf{55.1} & \textbf{28.4} & \textbf{62.0} & 35.5 & \textbf{52.6} & 59.2 & \textbf{46.8} & 70.0 & \textbf{65.2} & \textbf{61.7} & \underline{82.1} & 64.2 & \textbf{18.5} & \textbf{91.3} & \textbf{52.6} & \textbf{44.0} & \textbf{60.2}\\

\toprule
\multicolumn{21}{c}{\textbf{Clear-to-Adverse-Weather: Cityscapes$\to$ACDC (Test)}}  \\
\hline

ADVENT~\cite{vu2019advent} & 72.9 & 14.3 & 40.5 & 16.6 & 21.2 & 9.3 & 17.4 & 21.2 & 63.8 & 23.8 & 18.3 & 32.6 & 19.5 & 69.5 & 36.2 & 34.5 & 46.2 & 26.9 & 36.1 & 32.7\\
MGCDA$^\dagger$~\cite{sakaridis2020map} & 73.4 & 28.7 & 69.9 & 19.3 & 26.3 & 36.8 & 53.0 & 53.3 & \underline{75.4} & 32.0 & 84.6 & 51.0 & 26.1 & 77.6 & 43.2 & 45.9 & 53.9 & 32.7 & 41.5 & 48.7\\
DANNet$^\dagger$~\cite{wu2021dannet} & 84.3 & 54.2 & 77.6 & 38.0 & 30.0 & 18.9 & 41.6 & 35.2 & 71.3 & 39.4 & \underline{86.6} & 48.7 & 29.2 & 76.2 & 41.6 & 43.0 & 58.6 & 32.6 & 43.9 & 50.0\\
DAFormer~\cite{hoyer2021daformer} & 58.4 & 51.3 & 84.0 & 42.7 & 35.1 & 50.7 & 30.0 & 57.0 & 74.8 & 52.8 & 51.3 & 58.3 & 32.6 & 82.7 & 58.3 & 54.9 & 82.4 & 44.1 & 50.7 & 55.4\\
HRDA~\cite{hoyer2022hrda} & \underline{88.3} & \underline{57.9} & \underline{88.1} & \textbf{55.2} & \underline{36.7} & \underline{56.3} & \textbf{62.9} & \underline{65.3} & 74.2 & \underline{57.7} & 85.9 & \underline{68.8} & \underline{45.7} & \underline{88.5} & \textbf{76.4} & \underline{82.4} & \underline{87.7} & \underline{52.7} & \underline{60.4} & \underline{68.0}\\
MIC~(HRDA) & \textbf{90.8} & \textbf{67.1} & \textbf{89.2} & \underline{54.5} & \textbf{40.5} & \textbf{57.2} & \underline{62.0} & \textbf{68.4} & \textbf{76.3} & \textbf{61.8} & \textbf{87.0} & \textbf{71.3} & \textbf{49.4} & \textbf{89.7} & \underline{75.7} & \textbf{86.8} & \textbf{89.1} & \textbf{56.9} & \textbf{63.0} & \textbf{70.4}\\
\hline

\multicolumn{21}{l}{\rule{0pt}{3ex}$^\dagger$ Method uses additional daytime/clear-weather geographically-aligned reference images.}
\end{tabular}
\end{table*}

\begin{figure*}
\centering
{\footnotesize
\begin{tabularx}{\linewidth}{*{6}{Y}}
Image & 
DAFormer~\cite{hoyer2021daformer} & 
HRDA~\cite{hoyer2022hrda} & 
MIC (HRDA) & Ground Truth \\
\end{tabularx}
} %
\includegraphics[width=\linewidth]{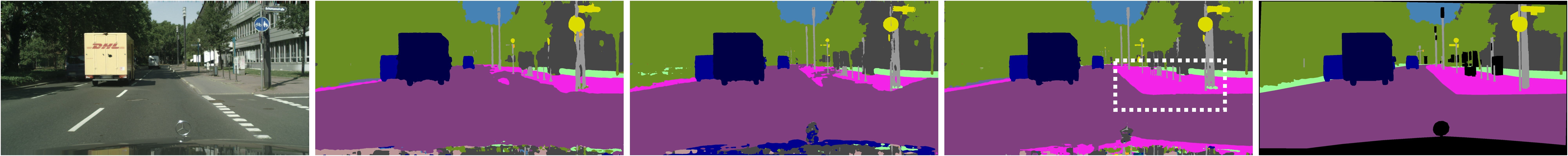}\\
\includegraphics[width=\linewidth]{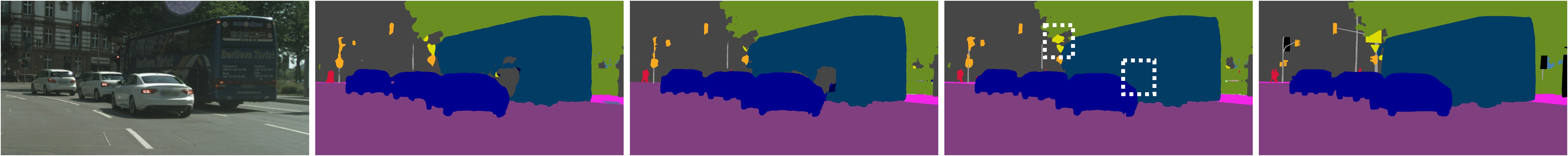}\\
\includegraphics[width=\linewidth]{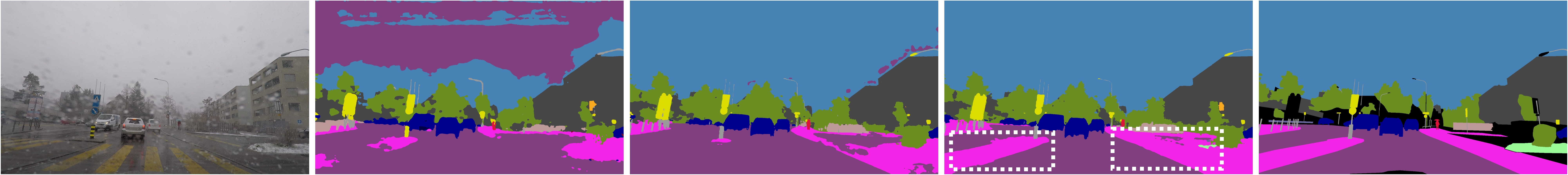}\\
\includegraphics[width=\linewidth]{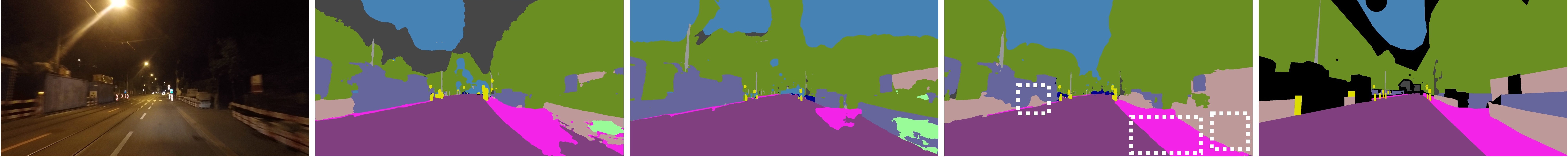}\\
\scriptsize%
\setlength\tabcolsep{1pt}%
{%
\newcolumntype{P}[1]{>{\centering\arraybackslash}p{#1}}
\begin{tabular}{@{}*{20}{P{0.09\columnwidth}}@{}}
     {\cellcolor[rgb]{0.5,0.25,0.5}}\textcolor{white}{road} 
     &{\cellcolor[rgb]{0.957,0.137,0.91}}sidew. 
     &{\cellcolor[rgb]{0.275,0.275,0.275}}\textcolor{white}{build.} 
     &{\cellcolor[rgb]{0.4,0.4,0.612}}\textcolor{white}{wall} 
     &{\cellcolor[rgb]{0.745,0.6,0.6}}fence 
     &{\cellcolor[rgb]{0.6,0.6,0.6}}pole 
     &{\cellcolor[rgb]{0.98,0.667,0.118}}tr. light
     &{\cellcolor[rgb]{0.863,0.863,0}}tr. sign 
     &{\cellcolor[rgb]{0.42,0.557,0.137}}veget. 
     &{\cellcolor[rgb]{0.596,0.984,0.596}}terrain 
     &{\cellcolor[rgb]{0.275,0.510,0.706}}sky
     &{\cellcolor[rgb]{0.863,0.078,0.235}}\textcolor{white}{person} 
     &{\cellcolor[rgb]{0.988,0.494,0.635}}\textcolor{black}{rider} 
     &{\cellcolor[rgb]{0,0,0.557}}\textcolor{white}{car} 
     &{\cellcolor[rgb]{0,0,0.275}}\textcolor{white}{truck} 
     &{\cellcolor[rgb]{0,0.235,0.392}}\textcolor{white}{bus}
     &{\cellcolor[rgb]{0,0.392,0.471}}\textcolor{white}{train} 
     &{\cellcolor[rgb]{0,0,0.902}}\textcolor{white}{m.bike} 
     & {\cellcolor[rgb]{0.467,0.043,0.125}}\textcolor{white}{bike}
     &{\cellcolor[rgb]{0,0,0}}\textcolor{white}{n/a.}
\end{tabular}
}%
\\
\caption{Qualitative comparison of MIC with previous methods on GTA$\to$CS (row 1 and 2), CS$\to$ACDC (row 3), and CS$\to$DarkZurich (row 4). MIC better segments difficult classes such as \emph{sidewalk}, \emph{fence}, \emph{traffic sign}, and \emph{bus}. Further examples are shown in the supplement.}
\label{fig:visual_examples}
\end{figure*}

\begin{table*}[ht!]
	\centering
	\caption{Image classification accuracy in \% on VisDA-2017 for UDA.
	The last column contains the mean across classes.
	}
	\scriptsize
	\setlength{\tabcolsep}{3pt}
	\label{table:visda}

	\begin{tabular}{l|c|cccccccccccc|c}
		\hline

		Method && Plane & Bcycl & Bus & Car & Horse & Knife & Mcyle & Persn & Plant & Sktb & Train & Truck & Mean\\
		\hline \hline
		CDAN \cite{long2018conditional} &\parbox[t]{2mm}{\multirow{4}{*}{\rotatebox[origin=c]{90}{ResNet}}}& 85.2 & 66.9 & 83.0 & 50.8 & 84.2 & 74.9 & 88.1 & 74.5 & 83.4 & 76.0 & 81.9 & 38.0 & 73.9\\
		MCC \cite{jin2020minimum} && 88.1 & 80.3 & \underline{80.5} & \underline{71.5} & 90.1 & 93.2 & 85.0 & 71.6 & 89.4 & 73.8 & 85.0 & 36.9 & 78.8\\
		SDAT \cite{rangwani2022closer} && \underline{95.8} & \underline{85.5} & 76.9 &69.0 & \underline{93.5} & \textbf{97.4} & \underline{88.5} & \underline{78.2} & \underline{93.1} & \underline{91.6} & \underline{86.3} & \underline{55.3} & \underline{84.3}\\
		MIC (SDAT) && \textbf{96.7} & \textbf{88.5} & \textbf{84.2} & \textbf{74.3} & \textbf{96.0} & \underline{96.3} & \textbf{90.2} & \textbf{81.2} & \textbf{94.3} & \textbf{95.4} & \textbf{88.9} & \textbf{56.6} & \textbf{86.9}\\
        \hline\hline
		TVT \cite{yang2021tvt} &\parbox[t]{2mm}{\multirow{5}{*}{\rotatebox[origin=c]{90}{ViT}}} & 92.9 & 85.6 & 77.5 & 60.5 & 93.6 & \underline{98.2} & 89.3 & 76.4 & 93.6 & 92.0 & 91.7 & 55.7 & 83.9 \\
		CDTrans \cite{xu2021cdtrans} && 97.1 & 90.5 & 82.4 & 77.5 & 96.6 & 96.1 & 93.6 & \underline{88.6} & \underline{97.9} & 86.9 & 90.3 & 62.8 & 88.4 \\
		SDAT \cite{rangwani2022closer} && \underline{98.4} & \underline{90.9} & \underline{85.4} & \underline{82.1} & \underline{98.5} & 97.6 & \underline{96.3} & 86.1 & 96.2 & \underline{96.7} & 92.9 & 56.8 & \underline{89.8} \\ 
 		SDAT w/ MAE~\cite{he2022masked} && 97.1 & 88.4 & 80.9 & 75.3 & 95.4 & 97.9 & 94.3 & 85.5 & 95.8 & 91.0 & \underline{93.0} & \underline{65.4} & 88.4 \\ 
		MIC (SDAT) && \textbf{99.0} & \textbf{93.3} & \textbf{86.5} & \textbf{87.6} & \textbf{98.9} & \textbf{99.0} & \textbf{97.2} & \textbf{89.8} & \textbf{98.9} & \textbf{98.9} & \textbf{96.5} & \textbf{68.0} & \textbf{92.8} \\

		\hline
	\end{tabular}%
\end{table*}

\begin{table}
  \centering     
  \caption{Image classification acc. in \% on Office-Home for UDA.}
  \setlength{\tabcolsep}{0.8pt}
  \scriptsize
  \begin{tabular}{l|cccccccccccc|c}
    \hline
    Method & A$\veryshortarrow$C & A$\veryshortarrow$P & A$\veryshortarrow$R & C$\veryshortarrow$A & C$\veryshortarrow$P & C$\veryshortarrow$R & P$\veryshortarrow$A & P$\veryshortarrow$C & P$\veryshortarrow$R & R$\veryshortarrow$A & R$\veryshortarrow$C & R$\veryshortarrow$P & Avg 
    \\\hline \hline

	CDTrans \cite{xu2021cdtrans} & 68.8 & 85.0 & 86.9 & 81.5 & 87.1 & 87.3 & 79.6 & 63.3 & 88.2 & 82.0 & 66.0 & 90.6 & 80.5 \\
	TVT \cite{yang2021tvt}  & \underline{74.9} & 86.8 & 89.5 & 82.8 & \underline{87.9} & 88.3 & 79.8 & \underline{71.9} & 90.1 & 85.5 & 74.6 & 90.6 & 83.6 \\
    SDAT \cite{rangwani2022closer} & 70.8 & \underline{87.0} & \underline{90.5} & \underline{85.2} & 87.3 & \underline{89.7} & \textbf{84.1} & 70.7 & \underline{90.6} & \underline{88.3} & \underline{75.5} & \textbf{92.1} & \underline{84.3} \\
    MIC (SDAT) & \textbf{80.2} & \textbf{87.3} & \textbf{91.1} & \textbf{87.2} & \textbf{90.0} & \textbf{90.1} & \underline{83.4} & \textbf{75.6}	& \textbf{91.2} & \textbf{88.6} & \textbf{78.7} & \underline{91.4} & \textbf{86.2}
    \\\hline
  \end{tabular}%

  \label{tab:officehome}
\end{table}
\begin{table}
	\centering
	\caption{Object detection AP in \% on CS$\to$Foggy CS.
	}
	\scriptsize
	\setlength{\tabcolsep}{2pt}
	\label{tab:dadetection1}
	\begin{tabular}{l|cccccccc|c}
		\hline

		Method                          & Bus  & Bcycl  & Car   &Mcycle & Persn & Rider & Train & Truck & mAP\\
		\hline \hline
		DAFaster \cite{chen2018domain} & 29.2 & 40.4   & 43.4  & 19.7  & 38.3  & 28.5& 23.7& 32.7& 32.0\\
		SW-DA \cite{saito2019strong}    & 31.8 & 44.3   &48.9   &21.0   &43.8   &28.0 &28.9 &\textbf{35.8} &35.3 \\
		SC-DA \cite{zhu2019adapting}       & 33.8 & 42.1   & 52.1  & 26.8  & 42.5  & 26.5 & 29.2 & \underline{34.5} &35.9 \\
		MTOR \cite{cai2019exploring}        & 38.6 & 35.6   & 44.0  & 28.3  & 30.6  & 41.4 & \underline{40.6} & 21.9 & 35.1 \\
		SIGMA \cite{li2022sigma} & \underline{50.4} & 40.6 & 60.3 & 31.7 & 44.0 & 43.9 & \textbf{51.5} & 31.6 & \underline{44.2} \\
		SADA \cite{chen2021scale}& 50.3 & \underline{45.4}   &\underline{62.1}   &\underline{32.4}   &\underline{48.5}   & \underline{52.6} &31.5 &29.5 &44.0 \\
MIC (SADA) & \textbf{52.4} & \textbf{47.5} & \textbf{67.0} & \textbf{40.6} & \textbf{50.9} & \textbf{55.3} & 33.7 & 33.9 & \textbf{47.6} \\
		\hline
	\end{tabular}
\end{table}

Second, we evaluate the performance of MIC combined with the best-performing UDA method HRDA~\cite{hoyer2022hrda} for further domain adaptation scenarios: synthetic-to-real (GTA$\to$\allowbreak CS and Synthia$\to$\allowbreak CS), day-to-nighttime (CS$\to$DarkZurich), and clear-to-adverse-weather (CS$\to$\allowbreak ACDC). Tab.~\ref{tab:sota_segmentation} shows clear performance improvements on each benchmark. Specifically, MIC improves the state-of-the-art performance by +2.1 mIoU on GTA$\to$CS, by +1.5 mIoU on Synthia$\to$CS, by +4.3 mIoU on CS$\to$\allowbreak DarkZurich, and by +2.4 mIoU on CS$\to$\allowbreak ACDC. 
Considering the class-wise IoU in Tab.~\ref{tab:sota_segmentation}, MIC achieves consistent improvements for most classes when compared to the previous state-of-the-art method HRDA. Classes that most profit from MIC are \emph{sidewalk}, \emph{fence}, \emph{pole}, \emph{traffic sign}, \emph{terrain}, and \emph{rider}. These classes have a comparably low UDA performance, meaning that they are difficult to adapt. Here, context clues appear to play an important role in successful adaptation.
For some classes such as \emph{building} or \emph{vegetation} on synthetic-to-real adaptation, MIC increases the performance by a smaller margin, probably because the target context clues play a smaller role for them.
In a few particular cases, the performance of single classes decreases for MIC such as \emph{truck} on CS$\to$DarkZurich. These are rare classes, which are underrepresented in the data, which might cause MIC to pick up misleading context biases.
The observations from Tab.~\ref{tab:sota_segmentation} are also reflected in the example predictions in Fig.~\ref{fig:visual_examples}. 
While previous methods often recognize only parts of ambiguous regions, MIC fixes these issues by using correctly detected parts as context. For instance, the grille of the bus in Fig.~\ref{fig:visual_examples} resembles a traffic cabinet (\emph{building}).
However, a cabinet between two vehicles is unlikely. 
Probably using this context prior, MIC can resolve the ambiguity.

\subsection{MIC for Image Classification}
\label{sec:exp_classification}

For image classification UDA, we combine MIC with the state-of-the-art method SDAT~\cite{rangwani2022closer}. On VisDA-2017 (Tab.~\ref{table:visda}), MIC significantly improves the UDA performance by +2.5 and +3.0 percent points when used with a ResNet and ViT network, respectively The improvement is consistent over almost all classes, where difficult classes generally benefit the most. On Office-Home (Tab.~\ref{tab:officehome}), MIC clearly improves the UDA performance by +1.9. Domains that are difficult to adapt such as A$\veryshortarrow$C or P$\veryshortarrow$C benefit most.
Tab.~\ref{table:visda} further provides a baseline of SDAT with MAE~\cite{he2022masked} pretraining, which includes masked image modeling (MIM) and ImageNet supervision. Compared to regular SDAT, additional MIM reduces the performance by -1.4. This demonstrates that naive MIM as additional pretraining is not sufficient to capture the relevant target context dependencies, probably as the learned context is specific to ImageNet and does not transfer well to the target domain.

\subsection{MIC for Object Detection}
For object detection UDA, we combine MIC with the state-of-the-art Scale-aware Domain Adaptive Faster-RCNN (SADA)~\cite{chen2021scale}.
On CS$\to$Foggy CS (Tab.~\ref{tab:dadetection1}), MIC obtains consistent improvements over all categories and achieves +3.6 mAP gain compared to the baseline SADA. The classes \emph{car}, \emph{motorcycle}, and \emph{rider} benefit the most. 
MIC also shows a clear advantage for most categories compared to more recent methods such as SIGMA~\cite{li2022sigma}.

\subsection{In-Depth Analysis of MIC}
\label{sec:exp_study}

\begin{figure}
    \centering
    \includegraphics[width=\linewidth]{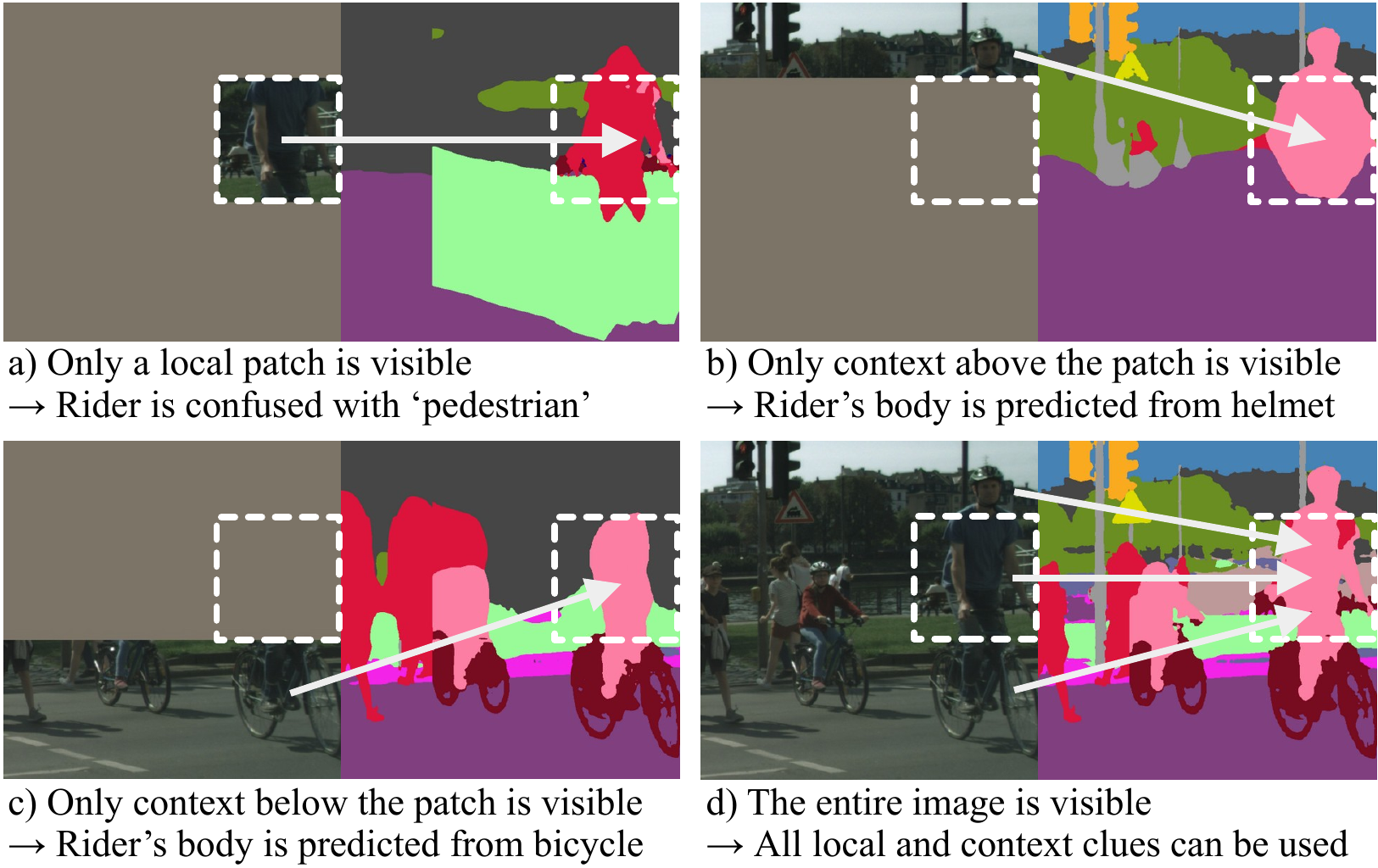}
    \caption{Predictions of MIC for masked variants of the same image demonstrating the learned context priors of MIC.
    }
    \label{fig:mic_context}
\end{figure}

\noindent\textbf{Context Utilization:}
To verify that MIC has learned context priors on the target domain, we mask out an image patch, let the trained model predict the semantics of the patch from the visible context, and calculate the mIoU for the patch. As the patch is masked out, 
the model can only utilize context information to infer its semantics. We repeat this process for all non-overlapping patches of the size $256{\times}256$ in the CS val. set. MIC(HRDA) achieves a strong context performance of 52.5 mIoU on GTA$\to$CS while HRDA only reaches 22.8 mIoU, showing that MIC indeed enhances context learning.

To further illustrate the learned
comprehensive context relations, we visualize predictions of
masked images in Fig.~\ref{fig:mic_context}. It shows the learned context
priors of helmet and bicycle, which MIC internally
exploits to predict the rider’s body.

\begin{table}[tb]
\centering
\caption{MIC with HRDA~\cite{hoyer2022hrda} for images from different domain.}
\label{tab:mic_domain}
\setlength{\tabcolsep}{5pt}
\scriptsize

\begin{tabular}{lcc}
\hline
   MIC Domain & $\text{mIoU}_\mathit{GTA\veryshortarrow CS}$ & $\text{mIoU}_\mathit{CS\veryshortarrow ACDC(Val)}$ \\
\hline\hline
           -- &    73.8 &             65.3 \\
       Source &    71.1 &             66.5 \\
       Target &    75.9 &             66.9 \\
Source+Target &    74.5 &             68.0 \\
\hline
\end{tabular}

\end{table}

\noindent\textbf{Where to apply MIC?}
Tab.~\ref{tab:mic_domain} shows the performance of MIC with HRDA using images from different domains as masked input: (1) source, (2) target, and (3) both source and target. We observe that: for (1) the performance is \mbox{-2.7} mIoU worse than HRDA for GTA$\to$CS but it increases by +1.2 mIoU for CS$\to$ACDC, for (2) the performance increases by +2.1 for GTA$\to$CS and +1.6 mIoU for CS$\to$ACDC, and for (3) the mIoU increases by +0.7 for GTA$\to$CS and +2.7 for CS$\to$ACDC. Both benchmarks differ in the domain gap of context relations. While the distributions of context relations can vary between synthetic (GTA) and real data (CS), the context relations of CS and ACDC are very similar as both datasets were recorded in the real world and partly even in the same city. If the context domain gap is large, context relations learned on source images do not transfer well to the target domain and can even hamper the adaptation. However, if the context gap is small, source context relations transfer well to the target domain and can boost the adaptation performance.
Therefore, we also apply MIC to the source domain, in addition to the default target domain, for clear-to-adverse-weather and day-to-nighttime adaptation.

\begin{table}[tb]
\centering
\caption{MIC ablation study with DAFormer~\cite{hoyer2021daformer} on GTA$\to$CS.}
\label{tab:mic_ablation}
\setlength{\tabcolsep}{3pt}
\scriptsize
\begin{tabular}{lccccc}
\hline
  & Masked Img. & Color Aug. & EMA Teacher & Pseudo Lbl. Weight & mIoU \\
\hline\hline
\arrayrulecolor{gray}
1 &          -- &         -- &          -- &                 -- & 68.3 \\
2 &         \cm &        \cm &         \cm &                \cm & 70.6 \\
\hline
3 &          -- &        \cm &         \cm &                \cm & 50.6 \\
4 &         \cm &         -- &         \cm &                \cm & 70.3 \\
5 &         \cm &        \cm &          -- &                \cm & 69.9 \\
6 &         \cm &        \cm &         \cm &                 -- & 69.0 \\
\arrayrulecolor{black}
\hline
\end{tabular}
\end{table}

\noindent\textbf{Component Ablation:}
To gain further insights, we ablate the components of MIC and evaluate the performance with DAFormer~\cite{hoyer2021daformer} (due to the faster training) on GTA$\to$CS in Tab.~\ref{tab:mic_ablation}. The complete MIC achieves 70.6 mIoU (row 2), which is +2.3 mIoU better than DAFormer (row 1).
First, the masking of the image is ablated, meaning that the consistency training is done with unmasked but still augmented target images (see ``MIC Parameters'' in Sec.~\ref{sec:implementation}). Without masking out image patches, the performance heavily decreases by -20.0 mIoU (cf. rows 2 and 3).
On the other side, color augmentation is not essential for MIC as its ablation only reduces the performance by -0.3 mIoU (cf. rows 2 and 4). This demonstrates the importance of context learning with masked images.
Replacing the EMA predictions with the regular model predictions decreases the performance of MIC by -0.7 mIoU (cf. rows 2 and 5). Without the pseudo-label confidence loss weight, the mIoU drops by -1.6 (cf. rows 2 and 6) showing that it is important to reduce the weight of uncertain samples for MIC training.

\begin{table}[tb]
\centering
\caption{Parameter study of the patch size $b$ and the mask ratio $r$ of MIC with DAFormer~\cite{hoyer2021daformer} on GTA$\to$CS. The color indicates the difference to the DAFormer performance of 68.3 mIoU.}
\label{fig:block_size_ratio_study}
\includegraphics[width=.85\linewidth]{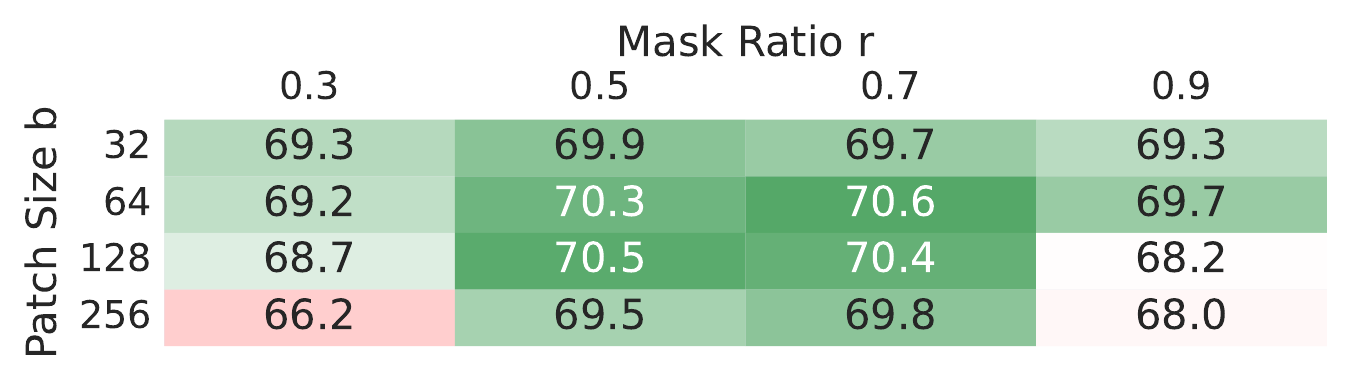}
\end{table}

\noindent\textbf{Patch Size and Mask Ratio:}
Tab.~\ref{fig:block_size_ratio_study} shows the influence of the mask patch size $b$ and mask ratio $r$. Compared to DAFormer, MIC achieves significant improvements in a range of $b$ between 64 and 128 and $r$ between 0.5 and 0.7. The best performance is achieved for $b{=}64$ and $r{=}0.7$. Only for a very large $b$ of 256, which is a quarter of the image height, MIC decreases the performance. Note that $b$ is internally divided by 2 as DAFormer uses half resolution. 

\noindent\textbf{MIC for Supervised Training:}
\begin{table}[tb]
\centering
\caption{Comparison of UDA on GTA$\to$CS and supervised training on CS. ``Rel.'' indicates $\text{mIoU}_\textit{UDA} / \text{mIoU}_\textit{Superv.}$.}
\label{tab:supervised}
\setlength{\tabcolsep}{5pt}
\scriptsize
\begin{tabular}{llll}
\hline
                 & $\text{mIoU}_\textit{UDA}$           & $\text{mIoU}_\textit{Superv.}$    & Rel.   \\
\hline\hline
\arrayrulecolor{gray}
DAFormer~\cite{hoyer2021daformer} & 68.3 & 77.6 & 88.0\%   \\
MIC (DAFormer)  & 70.6 & 77.9 & 90.6\%   \\
\hline
Improvement      & +2.3          & +0.3          & +2.6\% \\
\arrayrulecolor{black}
\hline
\end{tabular}
\end{table}

We compare the UDA and the supervised performance of DAFormer with and without MIC in Tab.~\ref{tab:supervised}. Also for supervised training, MIC achieves a slight improvement of +0.3 mIoU. However, the improvement for UDA is much more significant with +2.3 mIoU, showing that MIC is particularly useful for UDA. Therefore, MIC is able to increase the relative UDA performance (column ``Rel'') by +2.6 percent points, so that UDA with MIC achieves remarkable 90.6\% of the performance of a network trained with full supervision on the target domain.

\noindent\textbf{Runtime/Memory:}
\begin{table}
\centering
\caption{Runtime and memory consumption during training and inference on an RTX 2080 Ti (row 1-4) or Titan RTX (row 5-6).}
\label{tab:benchmark}
\setlength{\tabcolsep}{3pt}
\scriptsize
\begin{tabular}{lcccc}
\hline
                & \multicolumn{2}{c}{Training} & \multicolumn{2}{c}{Inference} \\
                & Throughput    & GPU Memory   & Throughput    & GPU Memory    \\
\hline\hline
\arrayrulecolor{gray}
Adversarial~\cite{tsai2018learning} & 1.40 it/s & 5.38 GB & 11.2 img/s & 0.5 GB \\
MIC (Adversarial) & 0.81 it/s & 5.55 GB & 11.2 img/s & 0.5 GB \\
\hline
DAFormer~\cite{hoyer2021daformer}        & 0.71 it/s      & 9.64 GB       & 8.6 img/s     & 1.0 GB \\
MIC (DAFormer) & 0.57 it/s      & 9.74 GB       & 8.6 img/s     & 1.0 GB  \\
\hline
HRDA~\cite{hoyer2022hrda} & 0.36 it/s & 22.46 GB & 0.8 img/s & 9.4 GB\\
MIC (HRDA) & 0.29 it/s & 22.55 GB & 0.8 img/s & 9.4 GB\\
\arrayrulecolor{black}
\hline
\end{tabular}
\end{table}
Tab.~\ref{tab:benchmark} shows the runtime and GPU memory footprint of representative UDA methods with and without MIC. For methods without an EMA teacher such as adversarial training, MIC reduces training speed by 75\% due to the additional calculations for MIC and increases the GPU memory consumption by 3\% due to the EMA teacher. The memory increase is small as the loss terms $\mathcal{L}^S$, $\mathcal{L}^T$, and $\mathcal{L}^M$ are backpropagated separately.
For UDA methods that already use an EMA teacher such as DAFormer or HRDA, the teacher and its predictions can be re-used, so that the training speed only increases by 24\% and the memory footprint by 1\%. 
Importantly, MIC is only used during training and does not increase the inference time at all.

\noindent\textbf{Supplement:} 
The supplement provides results on
CS$\to$\allowbreak Foggy\-Zurich,
further results of MIC with DAFormer and HRDA$_\text{DeepLabV2}$,
additional parameter and behavior studies,
an extended qualitative analysis, and further discussions.

\section{Conclusions}
\label{sec:conclusions}

In this paper, we presented Masked Image Consistency (MIC), a UDA module to improve the learning of target domain context relations. By enforcing consistency of predictions from partly masked and complete images, the network is trained to utilize robust context clues. 
MIC can be utilized for UDA across various visual recognition tasks such as image classification, semantic segmentation, and object detection as well as multiple domain adaptation scenarios such as synthetic-to-real, clear-to-adverse-weather, and day-to-nighttime. In a comprehensive evaluation, we have shown that MIC achieves significant performance improvements in all of these UDA tasks. For instance, MIC respectively improves the state-of-the-art performance by +2.1 and +3.0 on GTA$\to$CS and VisDA-2017. We hope that, due to its simplicity, MIC can be used as part of future UDA methods to narrow the gap between UDA and supervised learning.

\noindent\textbf{Acknowledgements:} This work is supported by the European Lighthouse on Secure and Safe AI (ELSA) and a Facebook Academic Gift on Robust Perception (INFO224).

{\small
\bibliographystyle{ieee_fullname}
\bibliography{literature}
}

\clearpage


\noindent\textbf{\Large Supplementary Material}

\makeatletter
\renewcommand{\theHsection}{papersection.\number\value{section}} 
\renewcommand{\thesection}{\Alph{section}}
\renewcommand{\thefigure}{S\arabic{figure}}
\renewcommand{\thetable}{S\arabic{table}}
\setcounter{section}{0}

\setcounter{figure}{0}
\setcounter{table}{0}
\makeatother

\section{Overview}

In the supplementary material for MIC, we provide the source code (Sec.~\ref{sec:supp_further_implementation_details}), study the influence of further MIC parameters (Sec.~\ref{sec:supp_parameters}), analyze additional aspects of the behavior of MIC (Sec.~\ref{sec:supp_extended_analysis}), extend the state-of-the-art comparison for semantic segmentation (Sec.~\ref{sec:supp_extended_sota}), provide a comprehensive qualitative comparison with previous works (Sec.~\ref{sec:supp_examples}), and discuss potential limitations (Sec.~\ref{sec:supp_limitations}).

\section{Source Code}
\label{sec:supp_further_implementation_details}

The source code to train MIC is available at \url{https://github.com/lhoyer/MIC}. For further information on the environment setup and experiment execution, please refer to \texttt{README.md}.
The implementation of MIC is based on the source code of HRDA~\cite{hoyer2022hrda} and mmsegmentation~\cite{mmseg2020} for semantic segmentation, SDAT~\cite{rangwani2022closer} for image classification, and SADA~\cite{chen2021scale} for object detection.

\section{Influence of Further MIC Parameters}
\label{sec:supp_parameters}

\subsection{MIC Prediction Region}

To gain a better understanding of the working principles of MIC, we additionally study how MIC behaves if only masked or unmasked regions of the image are included in the MIC loss (i.e. $\mathcal{L}^M$ is only calculated for regions, where $\mathcal{M}_{ij}$ is $0$ or $1$). Tab.~\ref{tab:mic_pred_region} shows that both MIC with a loss for masked patches and MIC with a loss for unmasked patches gain about +1.5 mIoU over DAFormer without MIC. When the MIC loss is calculated for both regions (default setting), the performance further improves by about +0.8 mIoU.

The improved performance for predicting masked patches shows that MIC profits from predicting regions with missing local information from the context. This task enhances the use of context relations for local predictions.

The improved performance for predicting unmasked patches shows that MIC profits from predicting regions with local information but without their complete context information. As not all context relations are available due to the masking, the network learns to exploit different combinations of context relations. This task enhances the robustness of the network towards missing context relations. During inference, this is particularly helpful to correctly predict partly-occluded objects (see Sec.~\ref{sec:supp_examples}).

Both capabilities are complementary and can be successfully combined when applying the MIC loss to all image patches.

\subsection{MIC Loss Weight $\lambda^M$}

Further, we study the influence of the MIC loss weight $\lambda^M$ with DAFormer on GTA$\to$CS. Tab.~\ref{tab:mic_lambda} shows that equal weighting of MIC loss ($\lambda^M=1$) and the other loss terms achieves the best performance. A smaller weight gradually degrades the performance up to the point where no MIC is used. Also, a larger loss weight results in a decreased performance. If it is too large such as $\lambda^M=10$, the performance can drop below the baseline. In that case, the MIC loss term dominates the total loss so that the other terms such as the source and adaptation loss cannot work effectively.

\begin{table}[tb]
\centering
\caption{Study of the MIC loss applied to specific image regions with DAFormer~\cite{hoyer2021daformer} on GTA$\to$CS.}
\label{tab:mic_pred_region}
\setlength{\tabcolsep}{3pt}
\scriptsize
\begin{tabular}{lc}
\toprule
MIC Loss Region & mIoU \\
\midrule
                   -- & 68.3 \\
        Masked Patches & 69.8 \\
        Unmasked Patches & 69.7 \\
            All Patches & 70.6 \\
\bottomrule
\end{tabular}
\end{table}

\begin{table}[tb]
\centering
\caption{Parameter study of the MIC loss weight $\lambda^M$ with DAFormer~\cite{hoyer2021daformer} on GTA$\to$CS.}
\label{tab:mic_lambda}
\setlength{\tabcolsep}{3pt}
\scriptsize
\begin{tabular}{cc}
\toprule
MIC Loss Weight $\lambda^M$ & mIoU \\
\midrule
\z0.0 & 68.3 \\
\z0.1 & 68.9 \\
\z0.5 & 69.5 \\
\z1.0 & 70.6 \\
\z2.0 & 70.1 \\
10.0 & 67.9 \\
\bottomrule
\end{tabular}
\end{table}

\begin{table}[tb]
\centering
\caption{Parameter study of the MIC teacher momentum $\alpha$ with DAFormer~\cite{hoyer2021daformer} on GTA$\to$CS and with SDAT~\cite{rangwani2022closer} on VisDA-2017.}
\label{tab:mic_alpha}
\setlength{\tabcolsep}{3pt}
\scriptsize
\begin{tabular}{ccc}
\toprule
Teacher Momentum $\alpha$ & $\text{mIoU}_\mathit{GTA\veryshortarrow CS}$ & $\text{mAcc}_\mathit{VisDA}$ \\
\midrule
\z\z\z0.9 & 70.0 & 92.8 \\
\z\z0.99 & 70.3 & 92.7 \\
\z0.999 & 70.6 & 80.5 \\
0.9999 & 69.3 & 79.5\\
\bottomrule
\end{tabular}
\end{table}

\begin{table}[tb]
\centering
\caption{Ablation study of color augmentation for MIC with DAFormer~\cite{hoyer2021daformer} on GTA$\to$CS and CS$\to$ACDC.}
\label{tab:mic_augmentation}
\setlength{\tabcolsep}{3pt}
\scriptsize
\begin{tabular}{lcc}
    \toprule
                MIC Domain & $\text{mIoU}_\mathit{GTA\veryshortarrow CS}$ & $\text{mIoU}_\mathit{CS\veryshortarrow ACDC(Val)}$ \\
    \midrule
    -- &    68.3 &     55.1 \\
    w/o Color Augmentation &    70.3 &     59.8 \\
     w/ Color Augmentation &    70.6 &     58.7 \\
    \bottomrule
    \end{tabular}
\end{table}

\subsection{Teacher Momentum $\alpha$}

Tab.~\ref{tab:mic_alpha} shows the influence of the MIC teacher network momentum $\alpha$ on the UDA performance for GTA$\to$Cityscapes (semantic segmentation) and VisDA-2017 (image classification). For GTA$\to$CS, it can be seen that the default value of $\alpha=0.999$ from DAFormer~\cite{hoyer2021daformer} achieves the best performance. A smaller $\alpha$ (faster teacher update) gradually decreases the performance. Similarly, a higher teacher $\alpha$ also results in a performance drop. Probably, a too large $\alpha$ (slow teacher update) results in outdated pseudo-labels, which hamper the consistency training. For VisDA-2017, $\alpha=0.9$ achieves the best performance, showing that a faster update of the teacher is useful for successful adaptation in this case.

\subsection{Data Augmentation on Different Datasets}

Tab.~\ref{tab:mic_augmentation} compares MIC without and with color augmentation (brightness, contrast, saturation, hue, and blur following the parameters of~\cite{tranheden2021dacs,hoyer2021daformer,hoyer2022hrda}) on GTA$\to$CS and CS$\to$ACDC. It can be seen that color augmentation improves MIC for GTA$\to$CS while it decreases the performance on CS$\to$ACDC. We assume that the color augmentation can corrupt the content of dark nighttime images due to the locally already low brightness and contrast. If the color augmentation corrupts the content of the unmasked patches of the image, the masked image consistency loss can be rendered meaningless. Therefore, we forgo color augmentation for target domains with nighttime images (DarkZurich and ACDC).

\begin{figure}
    \centering
    \includegraphics[width=\linewidth]{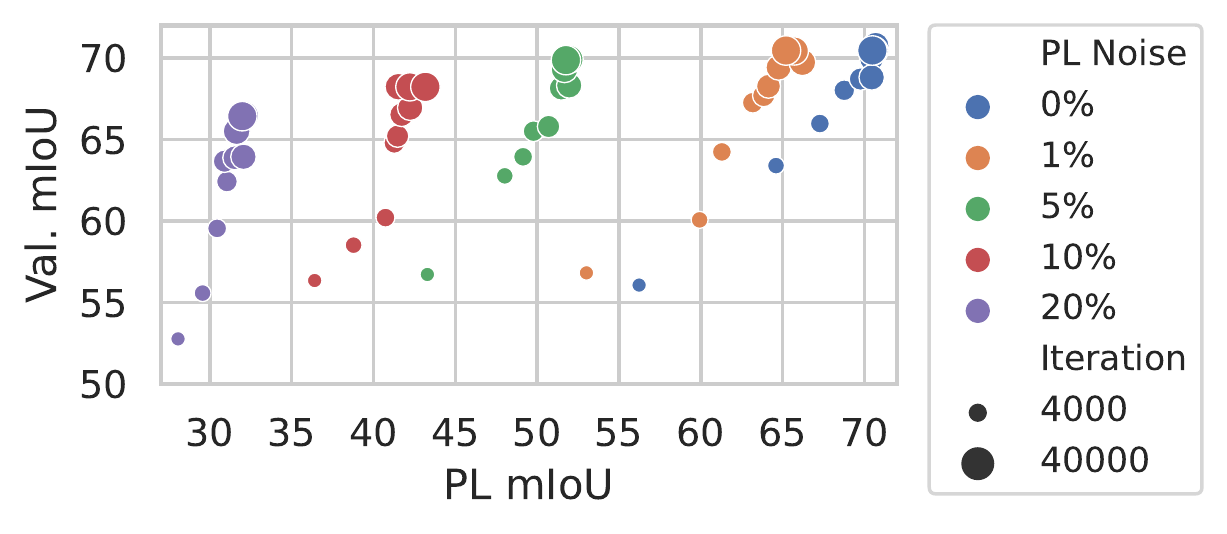}
    \caption{Influence of pseudo-label (PL) noise on the performance of MIC with DAFormer~\cite{hoyer2021daformer} on GTA$\rightarrow$CS through the training process at different training iterations.}
    \label{fig:pseudo_label_quality}
\end{figure}

\section{Extended Analysis of MIC}
\label{sec:supp_extended_analysis}

\subsection{Influence of Pseudo-Label Quality}
To analyze the influence of the pseudo-label (PL) quality on the performance of MIC through the training, Fig.~\ref{fig:pseudo_label_quality} plots the validation mIoU of MIC(DAFormer) with respect to the PL mIoU on the train set at several training iterations.
Both are clearly correlated, which is expected given that better PL improve the model and a better model improves PL by the EMA update.
To simulate worse PL, we add PL noise by randomly swapping the classes of PL segments. Even with 20\% PL noise, which reduces the PL mIoU by -38, the final val. mIoU only decreases by -4. This shows that MIC is relatively robust to PL noise during the training.

\subsection{MIC as Standalone}
MIC is designed as an orthogonal plug-in to enhance existing adaptation methods on various UDA benchmarks (see Tab.~1-5 in the main paper). Therefore, MIC requires an adaptation loss $\mathcal{L}^T$ from a host UDA method in order to work well. Without $\mathcal{L}^T$, the performance expectedly drops to 62.5 mIoU on GTA$\to$CS with a DAFormer network.
However, this is still +10.5 mIoU better than the host method in this case (DAFormer w/o $\mathcal{L}^T$).

\subsection{Why Selecting Random Patches?}
We have chosen random patch selection to promote a simple design that can be easily integrated into various UDA methods for different vision tasks and does not require a specific architecture or data priors.
Despite its simplicity, we show that it is a very powerful strategy (see Tab.~1-5 in the main paper).
As an image contains many objects with different context relations, it is hard to know in advance which relations are important. Random masking provides diverse context combinations through the training so that the network can identify different relevant relations.

\subsection{Why does MIC Work for Image Classification?}
Even though there is only one prediction per image for the image classification task, the network internally maintains spatial intermediate features. MIC can help to model context relations of object parts in these features, which can improve the distinction of ambiguous object parts on the target domain and reduce the effect of ill-adapted parts.

\section{Extended Comparison for UDA Semantic Segmentation}
\label{sec:supp_extended_sota}

\begin{table}
    \centering
    \caption{Semantic segmentation UDA on CS$\to$FoggyZurich}
    \label{tab:sota_foggyzurich}
    \setlength{\tabcolsep}{2.4pt}
    \scriptsize
    \begin{tabular}{lcc}
    \toprule
    Method                            & Training with Simulated Fog & mIoU \\
    \midrule
    CMAda2+~\cite{dai2020curriculum}  & \cm & 43.4 \\
    CMAda3+~\cite{dai2020curriculum}  & \cm & 46.8 \\
    FIFO~\cite{lee2022fifo}           & \cm & 48.4 \\
    CuDA-Net+~\cite{ma2022both}       & \cm & 49.1 \\
    \midrule
    DAFormer~\cite{hoyer2021daformer} & & 40.8 \\
    \textbf{MIC (DAFormer)}            & & 43.5 \\
    HRDA~\cite{hoyer2022hrda}         & & 46.0 \\
    \textbf{MIC (HRDA)}                & & \textbf{49.7} \\
    \bottomrule
    \end{tabular}
\end{table}

\begin{table*}
\centering
\caption{Extended comparison of the semantic segmentation performance (IoU in \%) on four different UDA benchmarks.}
\label{tab:extended_sota_segmentation}
\setlength{\tabcolsep}{2.4pt}
\scriptsize
\begin{tabular}{l|c|ccccccccccccccccccc|c}
\hline
Method & & Road & S.walk & Build. & Wall & Fence & Pole & Tr.Light & Sign & Veget. & Terrain & Sky & Person & Rider & Car & Truck & Bus & Train & M.bike & Bike & mIoU\\
\toprule
\multicolumn{21}{c}{\textbf{Synthetic-to-Real: GTA$\to$Cityscapes (Val.)}} \\
\hline

AdaptSeg~\cite{tsai2018learning} & \parbox[t]{2mm}{\multirow{15}{*}{\rotatebox[origin=c]{90}{ResNet-Based}}} & 86.5 & 25.9 & 79.8 & 22.1 & 20.0 & 23.6 & 33.1 & 21.8 & 81.8 & 25.9 & 75.9 & 57.3 & 26.2 & 76.3 & 29.8 & 32.1 & 7.2 & 29.5 & 32.5 & 41.4\\
ADVENT~\cite{vu2019advent} &  & 89.4 & 33.1 & 81.0 & 26.6 & 26.8 & 27.2 & 33.5 & 24.7 & 83.9 & 36.7 & 78.8 & 58.7 & 30.5 & 84.8 & 38.5 & 44.5 & 1.7 & 31.6 & 32.4 & 45.5\\
CBST~\cite{zou2018unsupervised} &  & 91.8 & 53.5 & 80.5 & 32.7 & 21.0 & 34.0 & 28.9 & 20.4 & 83.9 & 34.2 & 80.9 & 53.1 & 24.0 & 82.7 & 30.3 & 35.9 & 16.0 & 25.9 & 42.8 & 45.9\\
BDL~\cite{li2019bidirectional} &  & 91.0 & 44.7 & 84.2 & 34.6 & 27.6 & 30.2 & 36.0 & 36.0 & 85.0 & 43.6 & 83.0 & 58.6 & 31.6 & 83.3 & 35.3 & 49.7 & 3.3 & 28.8 & 35.6 & 48.5\\
FADA~\cite{wang2020classes} &  & 91.0 & 50.6 & 86.0 & 43.4 & 29.8 & 36.8 & 43.4 & 25.0 & 86.8 & 38.3 & 87.4 & 64.0 & 38.0 & 85.2 & 31.6 & 46.1 & 6.5 & 25.4 & 37.1 & 50.1\\
DACS~\cite{tranheden2021dacs} &  & 89.9 & 39.7 & 87.9 & 30.7 & 39.5 & 38.5 & 46.4 & 52.8 & 88.0 & 44.0 & 88.8 & 67.2 & 35.8 & 84.5 & 45.7 & 50.2 & 0.0 & 27.3 & 34.0 & 52.1\\
SAC~\cite{araslanov2021self} &  & 90.4 & 53.9 & 86.6 & 42.4 & 27.3 & 45.1 & 48.5 & 42.7 & 87.4 & 40.1 & 86.1 & 67.5 & 29.7 & 88.5 & 49.1 & 54.6 & 9.8 & 26.6 & 45.3 & 53.8\\
CorDA~\cite{wang2021domain} &  & 94.7 & 63.1 & 87.6 & 30.7 & 40.6 & 40.2 & 47.8 & 51.6 & 87.6 & 47.0 & 89.7 & 66.7 & 35.9 & 90.2 & 48.9 & 57.5 & 0.0 & 39.8 & 56.0 & 56.6\\
ProDA~\cite{zhang2021prototypical} &  & 87.8 & 56.0 & 79.7 & 46.3 & 44.8 & 45.6 & 53.5 & 53.5 & 88.6 & 45.2 & 82.1 & 70.7 & 39.2 & 88.8 & 45.5 & 59.4 & 1.0 & 48.9 & 56.4 & 57.5\\
ProCA~\cite{jiang2022prototypical} &  & 91.9 & 48.4 & 87.3 & 41.5 & 31.8 & 41.9 & 47.9 & 36.7 & 86.5 & 42.3 & 84.7 & 68.4 & 43.1 & 88.1 & 39.6 & 48.8 & 40.6 & 43.6 & 56.9 & 56.3\\
DecoupleNet~\cite{lai2022decouplenet} &  & 87.6 & 49.3 & 87.2 & 42.5 & 41.6 & 46.6 & 57.4 & 44.0 & 89.0 & 43.9 & 90.6 & 73.0 & 43.8 & 88.1 & 32.9 & 53.7 & 44.3 & 49.8 & 57.2 & 59.1\\
DAP~\cite{huo2022domain} &  & 94.5 & 63.1 & 89.1 & 29.8 & 47.5 & 50.4 & 56.7 & 58.7 & 89.5 & 50.2 & 87.0 & 73.6 & 38.6 & 91.3 & 50.2 & 52.9 & 0.0 & 50.2 & 63.5 & 59.8\\
CPSL~\cite{li2022class} &  & 92.3 & 59.9 & 84.9 & 45.7 & 29.7 & 52.8 & 61.5 & 59.5 & 87.9 & 41.6 & 85.0 & 73.0 & 35.5 & 90.4 & 48.7 & 73.9 & 26.3 & 53.8 & 53.9 & 60.8\\
HRDA$_\text{DLv2}$~\cite{hoyer2022hrda} &  & 96.2 & 73.1 & 89.7 & 43.2 & 39.9 & 47.5 & 60.0 & 60.0 & 89.9 & 47.1 & 90.2 & 75.9 & 49.0 & 91.8 & 61.9 & 59.3 & 10.2 & 47.0 & 65.3 & 63.0\\
\textbf{MIC~(HRDA$_\text{DLv2}$)} &  & 96.5 & 74.3 & 90.4 & 47.1 & 42.8 & 50.3 & 61.7 & 62.3 & 90.3 & 49.2 & 90.7 & 77.8 & \underline{53.2} & 93.0 & 66.2 & 68.0 & 6.8 & 38.0 & 60.6 & 64.2\\
\hline
DAFormer~\cite{hoyer2021daformer} & \parbox[t]{2mm}{\multirow{5}{*}{\rotatebox[origin=c]{90}{DAFormer}}} & 95.7 & 70.2 & 89.4 & 53.5 & 48.1 & 49.6 & 55.8 & 59.4 & 89.9 & 47.9 & 92.5 & 72.2 & 44.7 & 92.3 & 74.5 & 78.2 & 65.1 & 55.9 & 61.8 & 68.3\\
\textbf{MIC~(DAFormer)} &  & 96.7 & 75.0 & 90.0 & 58.2 & 50.4 & 51.1 & 56.7 & 62.1 & 90.2 & 51.3 & 92.9 & 72.4 & 47.1 & 92.8 & 78.9 & 83.4 & 75.6 & 54.2 & 62.6 & 70.6\\
\textbf{MIC~(DAFormer)$_\text{slide}$} &  & \underline{96.9} & \underline{76.5} & 90.1 & 57.6 & \underline{52.2} & 51.2 & 56.7 & 61.8 & 90.3 & \textbf{51.7} & 92.9 & 72.5 & 47.9 & 92.9 & 79.5 & 85.5 & \underline{76.8} & 53.6 & 62.9 & 71.0\\
HRDA~\cite{hoyer2022hrda} &  & 96.4 & 74.4 & \underline{91.0} & \textbf{61.6} & 51.5 & \underline{57.1} & \underline{63.9} & \underline{69.3} & \underline{91.3} & 48.4 & \underline{94.2} & \underline{79.0} & 52.9 & \underline{93.9} & \underline{84.1} & \underline{85.7} & 75.9 & \underline{63.9} & \underline{67.5} & \underline{73.8}\\
\textbf{MIC~(HRDA)} &  & \textbf{97.4} & \textbf{80.1} & \textbf{91.7} & \underline{61.2} & \textbf{56.9} & \textbf{59.7} & \textbf{66.0} & \textbf{71.3} & \textbf{91.7} & \underline{51.4} & \textbf{94.3} & \textbf{79.8} & \textbf{56.1} & \textbf{94.6} & \textbf{85.4} & \textbf{90.3} & \textbf{80.4} & \textbf{64.5} & \textbf{68.5} & \textbf{75.9}\\
\toprule
\multicolumn{21}{c}{\textbf{Synthetic-to-Real: Synthia$\to$Cityscapes (Val.)}} \\
\hline

ADVENT~\cite{vu2019advent} & \parbox[t]{2mm}{\multirow{13}{*}{\rotatebox[origin=c]{90}{ResNet-Based}}} & 85.6 & 42.2 & 79.7 & 8.7 & 0.4 & 25.9 & 5.4 & 8.1 & 80.4 & -- & 84.1 & 57.9 & 23.8 & 73.3 & -- & 36.4 & -- & 14.2 & 33.0 & 41.2\\
CBST~\cite{zou2018unsupervised} &  & 68.0 & 29.9 & 76.3 & 10.8 & 1.4 & 33.9 & 22.8 & 29.5 & 77.6 & -- & 78.3 & 60.6 & 28.3 & 81.6 & -- & 23.5 & -- & 18.8 & 39.8 & 42.6\\
FADA~\cite{wang2020classes} &  & 84.5 & 40.1 & 83.1 & 4.8 & 0.0 & 34.3 & 20.1 & 27.2 & 84.8 & -- & 84.0 & 53.5 & 22.6 & 85.4 & -- & 43.7 & -- & 26.8 & 27.8 & 45.2\\
DACS~\cite{tranheden2021dacs} &  & 80.6 & 25.1 & 81.9 & 21.5 & 2.9 & 37.2 & 22.7 & 24.0 & 83.7 & -- & 90.8 & 67.6 & 38.3 & 82.9 & -- & 38.9 & -- & 28.5 & 47.6 & 48.3\\
SAC~\cite{araslanov2021self} &  & 89.3 & 47.2 & 85.5 & 26.5 & 1.3 & 43.0 & 45.5 & 32.0 & 87.1 & -- & 89.3 & 63.6 & 25.4 & 86.9 & -- & 35.6 & -- & 30.4 & 53.0 & 52.6\\
CorDA~\cite{wang2021domain} &  & \textbf{93.3} & \textbf{61.6} & 85.3 & 19.6 & 5.1 & 37.8 & 36.6 & 42.8 & 84.9 & -- & 90.4 & 69.7 & 41.8 & 85.6 & -- & 38.4 & -- & 32.6 & 53.9 & 55.0\\
ProDA~\cite{zhang2021prototypical} &  & 87.8 & 45.7 & 84.6 & 37.1 & 0.6 & 44.0 & 54.6 & 37.0 & \underline{88.1} & -- & 84.4 & 74.2 & 24.3 & 88.2 & -- & 51.1 & -- & 40.5 & 45.6 & 55.5\\
ProCA~\cite{jiang2022prototypical} &  & \underline{90.5} & \underline{52.1} & 84.6 & 29.2 & 3.3 & 40.3 & 37.4 & 27.3 & 86.4 & -- & 85.9 & 69.8 & 28.7 & 88.7 & -- & 53.7 & -- & 14.8 & 54.8 & 53.0\\
DecoupleNet~\cite{lai2022decouplenet} &  & 77.8 & 48.6 & 75.6 & 32.0 & 1.9 & 44.4 & 52.9 & 38.5 & 87.8 & -- & 88.1 & 71.1 & 34.3 & 88.7 & -- & 58.8 & -- & 50.2 & 61.4 & 57.0\\
DAP~\cite{huo2022domain} &  & 84.2 & 46.5 & 82.5 & 35.1 & 0.2 & 46.7 & 53.6 & 45.7 & \textbf{89.3} & -- & 87.5 & 75.7 & 34.6 & \textbf{91.7} & -- & \textbf{73.5} & -- & 49.4 & 60.5 & 59.8\\
CPSL~\cite{li2022class} &  & 87.2 & 43.9 & 85.5 & 33.6 & 0.3 & 47.7 & 57.4 & 37.2 & 87.8 & -- & 88.5 & 79.0 & 32.0 & \underline{90.6} & -- & 49.4 & -- & 50.8 & 59.8 & 57.9\\
HRDA$_\text{DLv2}$~\cite{hoyer2022hrda} &  & 85.8 & 47.3 & 87.3 & 27.3 & 1.4 & 50.5 & 57.8 & \underline{61.0} & 87.4 & -- & 89.1 & 76.2 & 48.5 & 87.3 & -- & 49.3 & -- & 55.0 & \underline{68.2} & 61.2\\
\textbf{MIC~(HRDA$_\text{DLv2}$)} &  & 84.7 & 45.7 & 88.3 & 29.9 & 2.8 & 53.3 & 61.0 & 59.5 & 86.9 & -- & 88.8 & 78.2 & \underline{53.3} & 89.4 & -- & 58.8 & -- & 56.0 & \textbf{68.3} & 62.8\\
\hline
DAFormer~\cite{hoyer2021daformer} & \parbox[t]{2mm}{\multirow{5}{*}{\rotatebox[origin=c]{90}{DAFormer}}} & 84.5 & 40.7 & 88.4 & 41.5 & 6.5 & 50.0 & 55.0 & 54.6 & 86.0 & -- & 89.8 & 73.2 & 48.2 & 87.2 & -- & 53.2 & -- & 53.9 & 61.7 & 60.9\\
\textbf{MIC~(DAFormer)} &  & 83.0 & 40.9 & 88.2 & 37.6 & \textbf{9.0} & 52.4 & 56.0 & 56.5 & 87.6 & -- & \underline{93.4} & 74.2 & 51.4 & 87.1 & -- & 59.6 & -- & 57.9 & 61.2 & 62.2\\
\textbf{MIC~(DAFormer)$_\text{slide}$} &  & 82.6 & 40.7 & 88.3 & 40.2 & \textbf{9.0} & 52.4 & 55.7 & 56.6 & 87.6 & -- & \underline{93.4} & 74.1 & 52.5 & 87.2 & -- & 62.2 & -- & 57.4 & 61.1 & 62.6\\
HRDA~\cite{hoyer2022hrda} &  & 85.2 & 47.7 & \underline{88.8} & \textbf{49.5} & 4.8 & \underline{57.2} & \underline{65.7} & 60.9 & 85.3 & -- & 92.9 & \underline{79.4} & 52.8 & 89.0 & -- & \underline{64.7} & -- & \underline{63.9} & 64.9 & \underline{65.8}\\
\textbf{MIC~(HRDA)} &  & 86.6 & 50.5 & \textbf{89.3} & \underline{47.9} & 7.8 & \textbf{59.4} & \textbf{66.7} & \textbf{63.4} & 87.1 & -- & \textbf{94.6} & \textbf{81.0} & \textbf{58.9} & 90.1 & -- & 61.9 & -- & \textbf{67.1} & 64.3 & \textbf{67.3}\\

\toprule
\multicolumn{21}{c}{\textbf{Day-to-Nighttime: Cityscapes$\to$DarkZurich (Test)}} \\
\hline

ADVENT~\cite{vu2019advent} & \parbox[t]{2mm}{\multirow{10}{*}{\rotatebox[origin=c]{90}{ResNet-Based}}} & 85.8 & 37.9 & 55.5 & 27.7 & 14.5 & 23.1 & 14.0 & 21.1 & 32.1 & 8.7 & 2.0 & 39.9 & 16.6 & 64.0 & 13.8 & 0.0 & 58.8 & 28.5 & 20.7 & 29.7\\
AdaptSeg~\cite{tsai2018learning} &  & 86.1 & 44.2 & 55.1 & 22.2 & 4.8 & 21.1 & 5.6 & 16.7 & 37.2 & 8.4 & 1.2 & 35.9 & 26.7 & 68.2 & 45.1 & 0.0 & 50.1 & 33.9 & 15.6 & 30.4\\
BDL~\cite{li2019bidirectional} &  & 85.3 & 41.1 & 61.9 & 32.7 & 17.4 & 20.6 & 11.4 & 21.3 & 29.4 & 8.9 & 1.1 & 37.4 & 22.1 & 63.2 & 28.2 & 0.0 & 47.7 & 39.4 & 15.7 & 30.8\\
GCMA$^\dagger$~\cite{sakaridis2019guided} &  & 81.7 & 46.9 & 58.8 & 22.0 & 20.0 & 41.2 & 40.5 & 41.6 & 64.8 & 31.0 & 32.1 & 53.5 & 47.5 & 75.5 & 39.2 & 0.0 & 49.6 & 30.7 & 21.0 & 42.0\\
MGCDA$^\dagger$~\cite{sakaridis2020map} &  & 80.3 & 49.3 & 66.2 & 7.8 & 11.0 & 41.4 & 38.9 & 39.0 & 64.1 & 18.0 & 55.8 & 52.1 & 53.5 & 74.7 & 66.0 & 0.0 & 37.5 & 29.1 & 22.7 & 42.5\\
DANNet$^\dagger$~\cite{wu2021dannet} &  & 90.0 & 54.0 & 74.8 & 41.0 & 21.1 & 25.0 & 26.8 & 30.2 & \textbf{72.0} & 26.2 & \textbf{84.0} & 47.0 & 33.9 & 68.2 & 19.0 & 0.3 & 66.4 & 38.3 & 23.6 & 44.3\\
CDAda$^\dagger$~\cite{xu2021cdada} &  & 90.5 & 60.6 & 67.9 & 37.0 & 19.3 & 42.9 & 36.4 & 35.3 & 66.9 & 24.4 & 79.8 & 45.4 & 42.9 & 70.8 & 51.7 & 0.0 & 29.7 & 27.7 & 26.2 & 45.0\\
CCDistill$^\dagger$~\cite{gao2022cross} &  & 89.6 & 58.1 & 70.6 & 36.6 & 22.5 & 33.0 & 27.0 & 30.5 & 68.3 & 33.0 & \underline{80.9} & 42.3 & 40.1 & 69.4 & 58.1 & 0.1 & 72.6 & 47.7 & 21.3 & 47.5\\
HRDA$_\text{DLv2}$~\cite{hoyer2022hrda} &  & 88.7 & 65.5 & 68.3 & 41.9 & 18.1 & 50.6 & 6.0 & 39.6 & 33.3 & 34.4 & 0.3 & 57.6 & 51.7 & 75.0 & 70.9 & 8.5 & 63.6 & 41.0 & 38.8 & 44.9\\
\textbf{MIC~(HRDA$_\text{DLv2}$)} &  & 82.8 & \underline{69.6} & 75.5 & 44.0 & 21.0 & 51.1 & 43.4 & \underline{48.3} & 39.3 & 37.1 & 0.0 & 59.4 & 53.6 & 73.6 & \underline{74.2} & 9.2 & 78.7 & 40.0 & 37.2 & 49.4\\
\hline
DAFormer~\cite{hoyer2021daformer} & \parbox[t]{2mm}{\multirow{5}{*}{\rotatebox[origin=c]{90}{DAFormer}}} & \underline{93.5} & 65.5 & 73.3 & 39.4 & 19.2 & 53.3 & 44.1 & 44.0 & 59.5 & 34.5 & 66.6 & 53.4 & 52.7 & \underline{82.1} & 52.7 & 9.5 & 89.3 & 50.5 & 38.5 & 53.8\\
\textbf{MIC~(DAFormer)} &  & 88.2 & 60.5 & 73.5 & 53.5 & 23.8 & 52.3 & \underline{44.6} & 43.8 & 68.6 & 34.0 & 58.1 & 57.8 & 48.2 & 78.7 & 58.0 & 13.3 & 91.2 & 46.1 & 42.9 & 54.6\\
\textbf{MIC~(DAFormer)$_\text{slide}$} &  & 89.9 & 65.0 & \underline{75.9} & \underline{54.9} & \underline{25.5} & 53.3 & \underline{44.6} & 44.0 & \underline{70.0} & \underline{39.2} & 62.0 & 58.4 & 48.7 & 79.8 & 59.6 & \textbf{21.0} & \textbf{91.3} & \textbf{53.4} & \textbf{44.7} & \underline{56.9}\\
HRDA~\cite{hoyer2022hrda} &  & 90.4 & 56.3 & 72.0 & 39.5 & 19.5 & \underline{57.8} & \textbf{52.7} & 43.1 & 59.3 & 29.1 & 70.5 & \underline{60.0} & \underline{58.6} & \textbf{84.0} & \textbf{75.5} & 11.2 & 90.5 & 51.6 & 40.9 & 55.9\\
\textbf{MIC~(HRDA)} &  & \textbf{94.8} & \textbf{75.0} & \textbf{84.0} & \textbf{55.1} & \textbf{28.4} & \textbf{62.0} & 35.5 & \textbf{52.6} & 59.2 & \textbf{46.8} & 70.0 & \textbf{65.2} & \textbf{61.7} & \underline{82.1} & 64.2 & \underline{18.5} & \textbf{91.3} & \underline{52.6} & \underline{44.0} & \textbf{60.2}\\

\toprule
\multicolumn{21}{c}{\textbf{Clear-to-Adverse-Weather: Cityscapes$\to$ACDC (Test)}}  \\
\hline

ADVENT~\cite{vu2019advent} & \parbox[t]{2mm}{\multirow{9}{*}{\rotatebox[origin=c]{90}{ResNet-Based}}} & 72.9 & 14.3 & 40.5 & 16.6 & 21.2 & 9.3 & 17.4 & 21.2 & 63.8 & 23.8 & 18.3 & 32.6 & 19.5 & 69.5 & 36.2 & 34.5 & 46.2 & 26.9 & 36.1 & 32.7\\
AdaptSegNet~\cite{tsai2018learning} &  & 69.4 & 34.0 & 52.8 & 13.5 & 18.0 & 4.3 & 14.9 & 9.7 & 64.0 & 23.1 & 38.2 & 38.6 & 20.1 & 59.3 & 35.6 & 30.6 & 53.9 & 19.8 & 33.9 & 33.4\\
BDL~\cite{li2019bidirectional} &  & 56.0 & 32.5 & 68.1 & 20.1 & 17.4 & 15.8 & 30.2 & 28.7 & 59.9 & 25.3 & 37.7 & 28.7 & 25.5 & 70.2 & 39.6 & 40.5 & 52.7 & 29.2 & 38.4 & 37.7\\
GCMA$^\dagger$~\cite{sakaridis2019guided} &  & 79.7 & 48.7 & 71.5 & 21.6 & 29.9 & 42.5 & 56.7 & 57.7 & 75.8 & 39.5 & \underline{87.2} & 57.4 & 29.7 & 80.6 & 44.9 & 46.2 & 62.0 & 37.2 & 46.5 & 53.4\\
MGCDA$^\dagger$~\cite{sakaridis2020map} &  & 73.4 & 28.7 & 69.9 & 19.3 & 26.3 & 36.8 & 53.0 & 53.3 & 75.4 & 32.0 & 84.6 & 51.0 & 26.1 & 77.6 & 43.2 & 45.9 & 53.9 & 32.7 & 41.5 & 48.7\\
DANNet$^\dagger$~\cite{wu2021dannet} &  & 84.3 & 54.2 & 77.6 & 38.0 & 30.0 & 18.9 & 41.6 & 35.2 & 71.3 & 39.4 & 86.6 & 48.7 & 29.2 & 76.2 & 41.6 & 43.0 & 58.6 & 32.6 & 43.9 & 50.0\\
DANIA$^\dagger$~\cite{wu2021one} &  & 88.4 & 60.6 & 81.1 & 37.1 & 32.8 & 28.4 & 43.2 & 42.6 & \textbf{77.7} & 50.5 & \textbf{90.5} & 51.5 & 31.1 & 76.0 & 37.4 & 44.9 & 64.0 & 31.8 & 46.3 & 53.5\\
HRDA$_\text{DLv2}$~\cite{hoyer2022hrda} &  & 84.9 & 63.2 & 83.1 & 33.1 & 32.3 & 46.0 & 42.7 & 55.4 & 69.2 & 52.8 & 83.1 & 63.2 & 37.8 & 78.1 & 48.5 & 58.5 & 62.4 & 42.8 & 57.2 & 57.6\\
\textbf{MIC~(HRDA$_\text{DLv2}$)} &  & \underline{88.7} & \underline{63.9} & 84.1 & 38.4 & 35.7 & 45.7 & 51.5 & 60.3 & 72.7 & 52.3 & 85.8 & 62.5 & 39.8 & 84.7 & 37.7 & 68.7 & 71.9 & 46.0 & 56.5 & 60.4\\
\hline
DAFormer~\cite{hoyer2021daformer} & \parbox[t]{2mm}{\multirow{5}{*}{\rotatebox[origin=c]{90}{DAFormer}}} & 58.4 & 51.3 & 84.0 & 42.7 & 35.1 & 50.7 & 30.0 & 57.0 & 74.8 & 52.8 & 51.3 & 58.3 & 32.6 & 82.7 & 58.3 & 54.9 & 82.4 & 44.1 & 50.7 & 55.4\\
\textbf{MIC~(DAFormer)} &  & 58.5 & 51.6 & 84.9 & 48.1 & 39.8 & 50.8 & 39.7 & 59.9 & 77.1 & 54.9 & 51.9 & 63.9 & 40.7 & 84.1 & 63.1 & 66.2 & 85.5 & 46.3 & 57.1 & 59.2\\
\textbf{MIC~(DAFormer)$_\text{slide}$} &  & 60.5 & 60.5 & 86.1 & \underline{54.7} & \textbf{42.0} & 51.4 & 41.2 & 61.2 & \underline{77.6} & 57.4 & 53.6 & 64.6 & 40.2 & 85.9 & 68.7 & 73.8 & 87.0 & 50.1 & 58.8 & 61.9\\
HRDA~\cite{hoyer2022hrda} &  & 88.3 & 57.9 & \underline{88.1} & \textbf{55.2} & 36.7 & \underline{56.3} & \textbf{62.9} & \underline{65.3} & 74.2 & \underline{57.7} & 85.9 & \underline{68.8} & \underline{45.7} & \underline{88.5} & \textbf{76.4} & \underline{82.4} & \underline{87.7} & \underline{52.7} & \underline{60.4} & \underline{68.0}\\
\textbf{MIC~(HRDA)} &  & \textbf{90.8} & \textbf{67.1} & \textbf{89.2} & 54.5 & \underline{40.5} & \textbf{57.2} & \underline{62.0} & \textbf{68.4} & 76.3 & \textbf{61.8} & 87.0 & \textbf{71.3} & \textbf{49.4} & \textbf{89.7} & \underline{75.7} & \textbf{86.8} & \textbf{89.1} & \textbf{56.9} & \textbf{63.0} & \textbf{70.4}\\

\hline

\multicolumn{21}{l}{\rule{0pt}{3ex}$^\dagger$ Method uses additional daytime/clear-weather geographically-aligned reference images.}
\end{tabular}
\end{table*}

\paragraph{Cityscapes$\to$Foggy Zurich}
Supplementing the four semantic segmentation UDA benchmarks of the main paper, Tab.~\ref{tab:sota_foggyzurich} further provides the semantic segmentation performance of MIC on Cityscapes~\cite{cordts2016cityscapes} to Foggy Zurich~\cite{sakaridis2018model}. MIC was trained using the annotated Cityscapes training set as source domain and the unlabeled Foggy Zurich medium fog set as target domain. For validation, the model was tested on the Foggy Zurich test v2 set. Tab.~\ref{tab:sota_foggyzurich} shows that MIC(HRDA) significantly improves HRDA by +3.7 mIoU while MIC(DAFormer) gains +2.7 mIoU over DAFormer. MIC(HRDA) also outperforms specialized fog domain adaptation methods, which additionally utilize annotated Cityscapes images with simulated fog (Foggy Cityscapes DBF~\cite{sakaridis2018model}) during training.

\paragraph{Additional Baselines}
In the main paper, we have shown a selection of the most relevant methods for domain-adaptive semantic segmentation. In the extended comparison in Tab.~\ref{tab:extended_sota_segmentation}, we supplement the selection of previous works. It can be observed that also in the extended comparison, MIC(HRDA) outperforms all previous methods by a large margin. There are a few cases, where another method achieves a better performance for a specific class (e.g. DAP~\cite{huo2022domain} for vegetation on Synthia$\to$Cityscapes) but their performance falls behind MIC for other classes, resulting in a lower mIoU.

\paragraph{MIC with DAFormer}
Further, we provide MIC with DAFormer on all four benchmarks in Tab.~\ref{tab:extended_sota_segmentation}. Compared to DAFormer, MIC(DAFormer) achieves significant performance improvements across the different datasets. The performance of MIC(DAFormer) can be further improved by utilizing sliding window inference as suggested in HRDA~\cite{hoyer2022hrda} to use the same inference input size as the training crop, which works better for the learned positional embedding of the Transformer encoder. MIC(DAFormer)$_\text{slide}$ improves the performance on all four benchmarks, especially for day-to-nighttime and clear-to-adverse-weather adaptation.
Similar to MIC(HRDA), major improvements come from the classes \emph{sidewalk}, \emph{fence}, \emph{pole}, \emph{traffic sign}, \emph{terrain}, and \emph{rider}.

\paragraph{MIC with DeepLabV2}
For a more fair comparison with ResNet-based UDA methods, we further provide detailed results of MIC(HRDA$_\text{DLv2}$), which uses a DeepLabV2~\cite{chen2017deeplab} network architecture with a ResNet-101~\cite{he2016deep} backbone, in Tab.~\ref{tab:extended_sota_segmentation}. It can be seen that MIC(HRDA$_\text{DLv2}$) significantly outperforms recent ResNet-based methods such as DecoupleNet~\cite{lai2022decouplenet}, DAP~\cite{huo2022domain}, CPSL~\cite{li2022class}, and HRDA$_\text{DLv2}$~\cite{hoyer2022hrda} on synthetic-to-real adaptation as well as CCDistill~\cite{gao2022cross}, DANIA~\cite{wu2021one}, and HRDA$_\text{DLv2}$~\cite{hoyer2022hrda} on day-to-nighttime/clear-to-adverse-weather adaptation.

\section{Further Example Predictions}
\label{sec:supp_examples}

\begin{figure*}
\centering
{\footnotesize
\begin{tabularx}{\linewidth}{*{7}{Y}}
Image & 
ProDA~\cite{zhang2021prototypical} & 
DAFormer~\cite{hoyer2021daformer} & 
HRDA~\cite{hoyer2022hrda} & 
MIC (HRDA) & Ground Truth \\
\end{tabularx}
} %
\includegraphics[width=\linewidth]{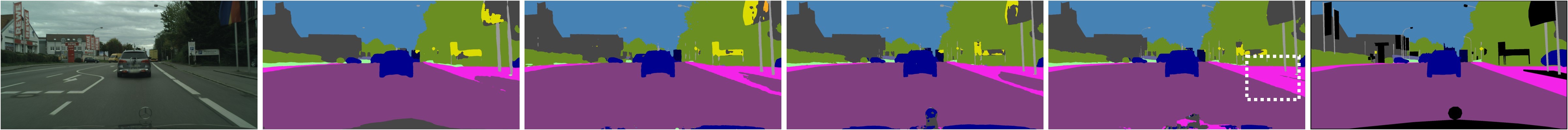}
\includegraphics[width=\linewidth]{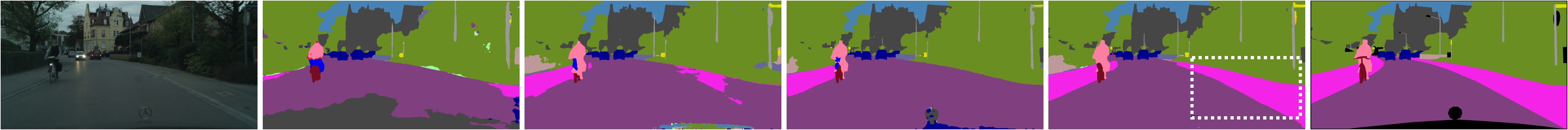}
\includegraphics[width=\linewidth]{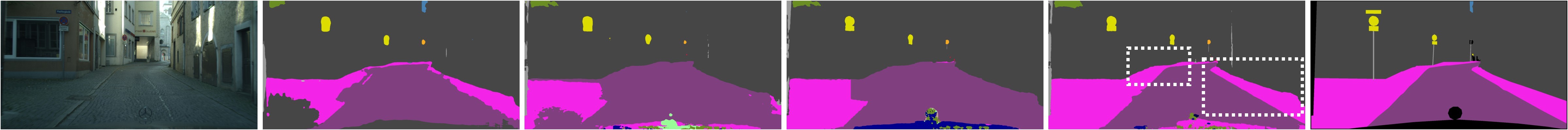}
\scriptsize%
\setlength\tabcolsep{1pt}%
{%
\newcolumntype{P}[1]{>{\centering\arraybackslash}p{#1}}
\begin{tabular}{@{}*{20}{P{0.09\columnwidth}}@{}}
     {\cellcolor[rgb]{0.5,0.25,0.5}}\textcolor{white}{road} 
     &{\cellcolor[rgb]{0.957,0.137,0.91}}sidew. 
     &{\cellcolor[rgb]{0.275,0.275,0.275}}\textcolor{white}{build.} 
     &{\cellcolor[rgb]{0.4,0.4,0.612}}\textcolor{white}{wall} 
     &{\cellcolor[rgb]{0.745,0.6,0.6}}fence 
     &{\cellcolor[rgb]{0.6,0.6,0.6}}pole 
     &{\cellcolor[rgb]{0.98,0.667,0.118}}tr. light
     &{\cellcolor[rgb]{0.863,0.863,0}}tr. sign 
     &{\cellcolor[rgb]{0.42,0.557,0.137}}veget. 
     &{\cellcolor[rgb]{0.596,0.984,0.596}}terrain 
     &{\cellcolor[rgb]{0.275,0.510,0.706}}sky
     &{\cellcolor[rgb]{0.863,0.078,0.235}}\textcolor{white}{person} 
     &{\cellcolor[rgb]{0.988,0.494,0.635}}\textcolor{black}{rider} 
     &{\cellcolor[rgb]{0,0,0.557}}\textcolor{white}{car} 
     &{\cellcolor[rgb]{0,0,0.275}}\textcolor{white}{truck} 
     &{\cellcolor[rgb]{0,0.235,0.392}}\textcolor{white}{bus}
     &{\cellcolor[rgb]{0,0.392,0.471}}\textcolor{white}{train} 
     &{\cellcolor[rgb]{0,0,0.902}}\textcolor{white}{m.bike} 
     & {\cellcolor[rgb]{0.467,0.043,0.125}}\textcolor{white}{bike}
     &{\cellcolor[rgb]{0,0,0}}\textcolor{white}{n/a.}
\end{tabular}
}%

\vspace{-0.15cm}
\caption{Example predictions showing a better segmentation of \emph{sidewalk} by MIC on GTA$\rightarrow$Cityscapes.}
\label{fig:predictions_gta_sidewalk}
\end{figure*}

\begin{figure*}
\centering
{\footnotesize
\begin{tabularx}{\linewidth}{*{7}{Y}}
Image & 
ProDA~\cite{zhang2021prototypical} & 
DAFormer~\cite{hoyer2021daformer} & 
HRDA~\cite{hoyer2022hrda} & 
MIC (HRDA) & Ground Truth \\
\end{tabularx}
} %
\includegraphics[width=\linewidth]{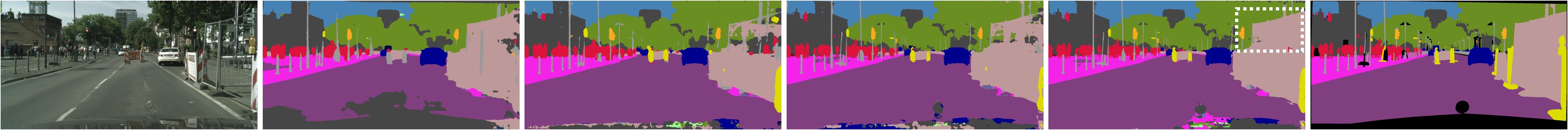}
\includegraphics[width=\linewidth]{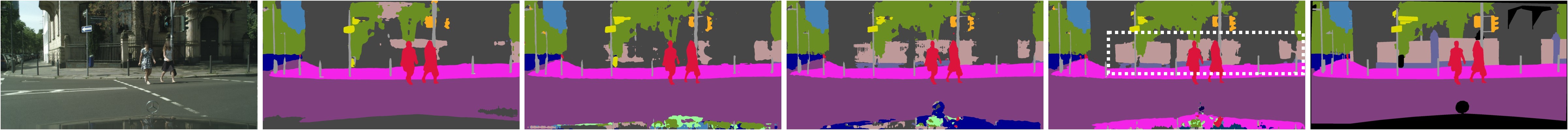}
\includegraphics[width=\linewidth]{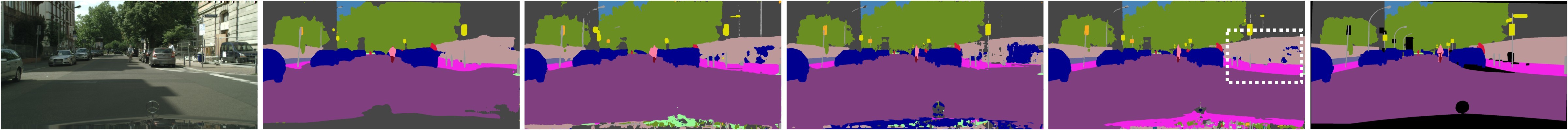}
\scriptsize%
\setlength\tabcolsep{1pt}%
{%
\newcolumntype{P}[1]{>{\centering\arraybackslash}p{#1}}
\begin{tabular}{@{}*{20}{P{0.09\columnwidth}}@{}}
     {\cellcolor[rgb]{0.5,0.25,0.5}}\textcolor{white}{road} 
     &{\cellcolor[rgb]{0.957,0.137,0.91}}sidew. 
     &{\cellcolor[rgb]{0.275,0.275,0.275}}\textcolor{white}{build.} 
     &{\cellcolor[rgb]{0.4,0.4,0.612}}\textcolor{white}{wall} 
     &{\cellcolor[rgb]{0.745,0.6,0.6}}fence 
     &{\cellcolor[rgb]{0.6,0.6,0.6}}pole 
     &{\cellcolor[rgb]{0.98,0.667,0.118}}tr. light
     &{\cellcolor[rgb]{0.863,0.863,0}}tr. sign 
     &{\cellcolor[rgb]{0.42,0.557,0.137}}veget. 
     &{\cellcolor[rgb]{0.596,0.984,0.596}}terrain 
     &{\cellcolor[rgb]{0.275,0.510,0.706}}sky
     &{\cellcolor[rgb]{0.863,0.078,0.235}}\textcolor{white}{person} 
     &{\cellcolor[rgb]{0.988,0.494,0.635}}\textcolor{black}{rider} 
     &{\cellcolor[rgb]{0,0,0.557}}\textcolor{white}{car} 
     &{\cellcolor[rgb]{0,0,0.275}}\textcolor{white}{truck} 
     &{\cellcolor[rgb]{0,0.235,0.392}}\textcolor{white}{bus}
     &{\cellcolor[rgb]{0,0.392,0.471}}\textcolor{white}{train} 
     &{\cellcolor[rgb]{0,0,0.902}}\textcolor{white}{m.bike} 
     & {\cellcolor[rgb]{0.467,0.043,0.125}}\textcolor{white}{bike}
     &{\cellcolor[rgb]{0,0,0}}\textcolor{white}{n/a.}
\end{tabular}
}%

\vspace{-0.15cm}
\caption{Example predictions showing a better segmentation of \emph{fence} by MIC on GTA$\rightarrow$Cityscapes.}
\label{fig:predictions_gta_fence}
\end{figure*}

\begin{figure*}
\centering
{\footnotesize
\begin{tabularx}{\linewidth}{*{7}{Y}}
Image & 
ProDA~\cite{zhang2021prototypical} & 
DAFormer~\cite{hoyer2021daformer} & 
HRDA~\cite{hoyer2022hrda} & 
MIC (HRDA) & Ground Truth \\
\end{tabularx}
} %
\includegraphics[width=\linewidth]{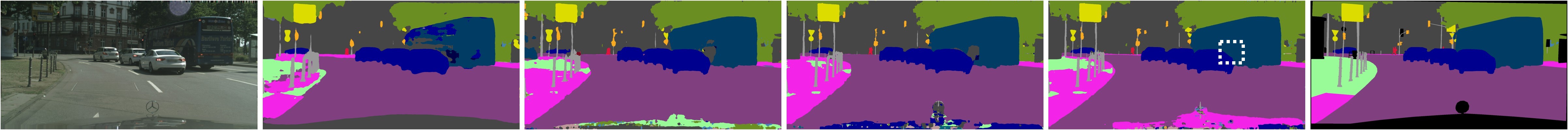}
\includegraphics[width=\linewidth]{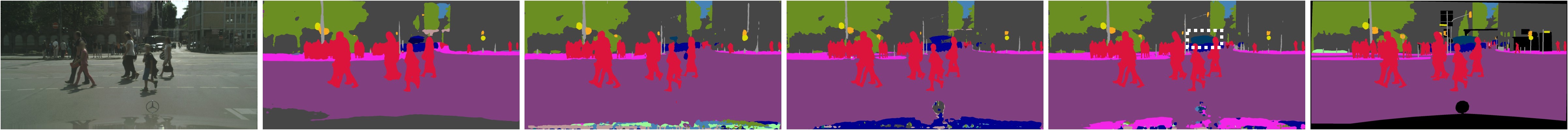}
\includegraphics[width=\linewidth]{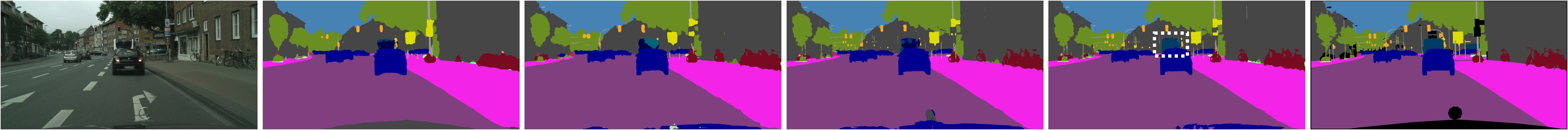}
\scriptsize%
\setlength\tabcolsep{1pt}%
{%
\newcolumntype{P}[1]{>{\centering\arraybackslash}p{#1}}
\begin{tabular}{@{}*{20}{P{0.09\columnwidth}}@{}}
     {\cellcolor[rgb]{0.5,0.25,0.5}}\textcolor{white}{road} 
     &{\cellcolor[rgb]{0.957,0.137,0.91}}sidew. 
     &{\cellcolor[rgb]{0.275,0.275,0.275}}\textcolor{white}{build.} 
     &{\cellcolor[rgb]{0.4,0.4,0.612}}\textcolor{white}{wall} 
     &{\cellcolor[rgb]{0.745,0.6,0.6}}fence 
     &{\cellcolor[rgb]{0.6,0.6,0.6}}pole 
     &{\cellcolor[rgb]{0.98,0.667,0.118}}tr. light
     &{\cellcolor[rgb]{0.863,0.863,0}}tr. sign 
     &{\cellcolor[rgb]{0.42,0.557,0.137}}veget. 
     &{\cellcolor[rgb]{0.596,0.984,0.596}}terrain 
     &{\cellcolor[rgb]{0.275,0.510,0.706}}sky
     &{\cellcolor[rgb]{0.863,0.078,0.235}}\textcolor{white}{person} 
     &{\cellcolor[rgb]{0.988,0.494,0.635}}\textcolor{black}{rider} 
     &{\cellcolor[rgb]{0,0,0.557}}\textcolor{white}{car} 
     &{\cellcolor[rgb]{0,0,0.275}}\textcolor{white}{truck} 
     &{\cellcolor[rgb]{0,0.235,0.392}}\textcolor{white}{bus}
     &{\cellcolor[rgb]{0,0.392,0.471}}\textcolor{white}{train} 
     &{\cellcolor[rgb]{0,0,0.902}}\textcolor{white}{m.bike} 
     & {\cellcolor[rgb]{0.467,0.043,0.125}}\textcolor{white}{bike}
     &{\cellcolor[rgb]{0,0,0}}\textcolor{white}{n/a.}
\end{tabular}
}%

\vspace{-0.15cm}
\caption{Example predictions showing a better segmentation of \emph{bus} by MIC on GTA$\rightarrow$Cityscapes.}
\label{fig:predictions_gta_bus}
\end{figure*}

\begin{figure*}
\centering
{\footnotesize
\begin{tabularx}{\linewidth}{*{7}{Y}}
Image & 
ProDA~\cite{zhang2021prototypical} & 
DAFormer~\cite{hoyer2021daformer} & 
HRDA~\cite{hoyer2022hrda} & 
MIC (HRDA) & Ground Truth \\
\end{tabularx}
} %
\includegraphics[width=\linewidth]{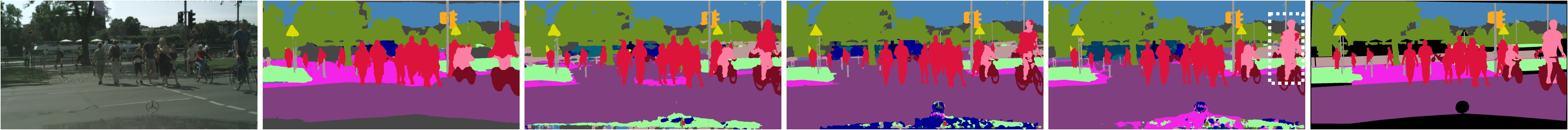}
\includegraphics[width=\linewidth]{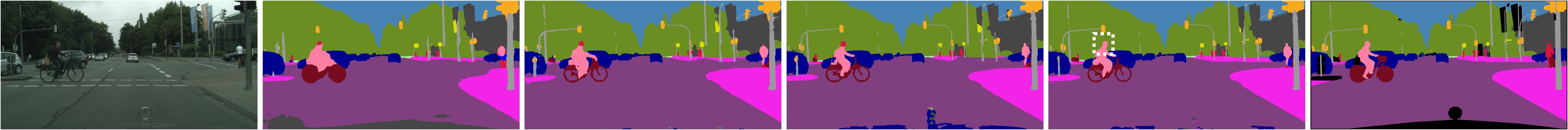}
\includegraphics[width=\linewidth]{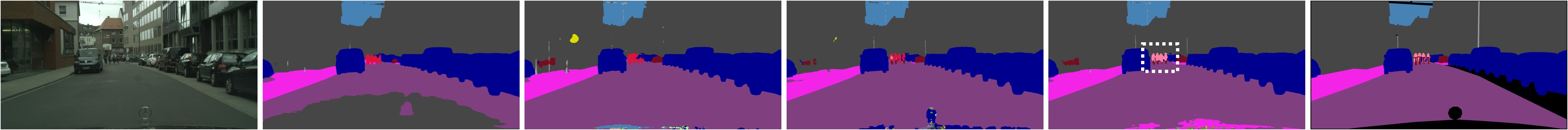}
\scriptsize%
\setlength\tabcolsep{1pt}%
{%
\newcolumntype{P}[1]{>{\centering\arraybackslash}p{#1}}
\begin{tabular}{@{}*{20}{P{0.09\columnwidth}}@{}}
     {\cellcolor[rgb]{0.5,0.25,0.5}}\textcolor{white}{road} 
     &{\cellcolor[rgb]{0.957,0.137,0.91}}sidew. 
     &{\cellcolor[rgb]{0.275,0.275,0.275}}\textcolor{white}{build.} 
     &{\cellcolor[rgb]{0.4,0.4,0.612}}\textcolor{white}{wall} 
     &{\cellcolor[rgb]{0.745,0.6,0.6}}fence 
     &{\cellcolor[rgb]{0.6,0.6,0.6}}pole 
     &{\cellcolor[rgb]{0.98,0.667,0.118}}tr. light
     &{\cellcolor[rgb]{0.863,0.863,0}}tr. sign 
     &{\cellcolor[rgb]{0.42,0.557,0.137}}veget. 
     &{\cellcolor[rgb]{0.596,0.984,0.596}}terrain 
     &{\cellcolor[rgb]{0.275,0.510,0.706}}sky
     &{\cellcolor[rgb]{0.863,0.078,0.235}}\textcolor{white}{person} 
     &{\cellcolor[rgb]{0.988,0.494,0.635}}\textcolor{black}{rider} 
     &{\cellcolor[rgb]{0,0,0.557}}\textcolor{white}{car} 
     &{\cellcolor[rgb]{0,0,0.275}}\textcolor{white}{truck} 
     &{\cellcolor[rgb]{0,0.235,0.392}}\textcolor{white}{bus}
     &{\cellcolor[rgb]{0,0.392,0.471}}\textcolor{white}{train} 
     &{\cellcolor[rgb]{0,0,0.902}}\textcolor{white}{m.bike} 
     & {\cellcolor[rgb]{0.467,0.043,0.125}}\textcolor{white}{bike}
     &{\cellcolor[rgb]{0,0,0}}\textcolor{white}{n/a.}
\end{tabular}
}%

\vspace{-0.15cm}
\caption{Example predictions showing a better segmentation of \emph{rider} by MIC on GTA$\rightarrow$Cityscapes.}
\label{fig:predictions_gta_rider}
\end{figure*}

\begin{figure*}
\centering
{\footnotesize
\begin{tabularx}{\linewidth}{*{7}{Y}}
Image & 
ProDA~\cite{zhang2021prototypical} & 
DAFormer~\cite{hoyer2021daformer} & 
HRDA~\cite{hoyer2022hrda} & 
MIC (HRDA) & Ground Truth \\
\end{tabularx}
} %
\includegraphics[width=\linewidth]{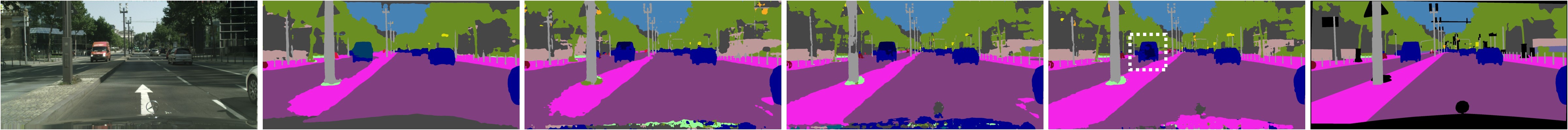}
\includegraphics[width=\linewidth]{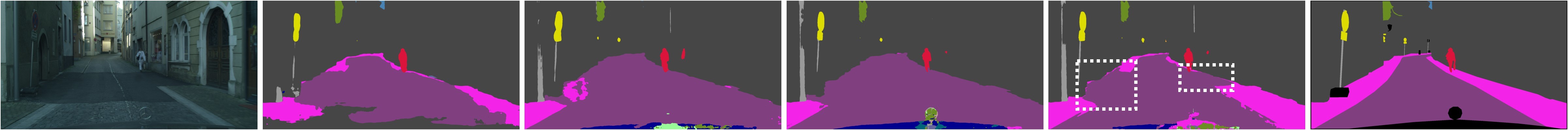}
\includegraphics[width=\linewidth]{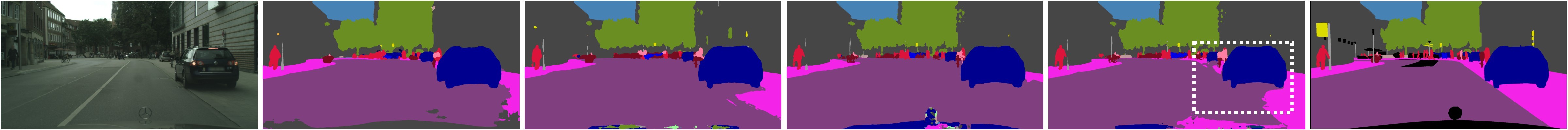}
\includegraphics[width=\linewidth]{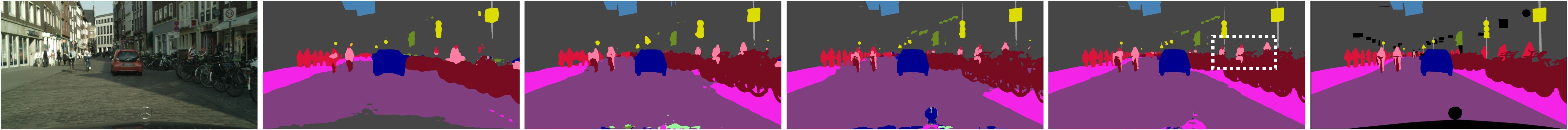}
\scriptsize%
\setlength\tabcolsep{1pt}%
{%
\newcolumntype{P}[1]{>{\centering\arraybackslash}p{#1}}
\begin{tabular}{@{}*{20}{P{0.09\columnwidth}}@{}}
     {\cellcolor[rgb]{0.5,0.25,0.5}}\textcolor{white}{road} 
     &{\cellcolor[rgb]{0.957,0.137,0.91}}sidew. 
     &{\cellcolor[rgb]{0.275,0.275,0.275}}\textcolor{white}{build.} 
     &{\cellcolor[rgb]{0.4,0.4,0.612}}\textcolor{white}{wall} 
     &{\cellcolor[rgb]{0.745,0.6,0.6}}fence 
     &{\cellcolor[rgb]{0.6,0.6,0.6}}pole 
     &{\cellcolor[rgb]{0.98,0.667,0.118}}tr. light
     &{\cellcolor[rgb]{0.863,0.863,0}}tr. sign 
     &{\cellcolor[rgb]{0.42,0.557,0.137}}veget. 
     &{\cellcolor[rgb]{0.596,0.984,0.596}}terrain 
     &{\cellcolor[rgb]{0.275,0.510,0.706}}sky
     &{\cellcolor[rgb]{0.863,0.078,0.235}}\textcolor{white}{person} 
     &{\cellcolor[rgb]{0.988,0.494,0.635}}\textcolor{black}{rider} 
     &{\cellcolor[rgb]{0,0,0.557}}\textcolor{white}{car} 
     &{\cellcolor[rgb]{0,0,0.275}}\textcolor{white}{truck} 
     &{\cellcolor[rgb]{0,0.235,0.392}}\textcolor{white}{bus}
     &{\cellcolor[rgb]{0,0.392,0.471}}\textcolor{white}{train} 
     &{\cellcolor[rgb]{0,0,0.902}}\textcolor{white}{m.bike} 
     & {\cellcolor[rgb]{0.467,0.043,0.125}}\textcolor{white}{bike}
     &{\cellcolor[rgb]{0,0,0}}\textcolor{white}{n/a.}
\end{tabular}
}%

\vspace{-0.15cm}
\caption{Failure cases of MIC on GTA$\rightarrow$Cityscapes.}
\label{fig:predictions_gta_failure}
\end{figure*}

\begin{figure*}
\centering
{\footnotesize
\begin{tabularx}{\linewidth}{*{6}{Y}}
Image & 
DAFormer~\cite{hoyer2021daformer} & 
HRDA~\cite{hoyer2022hrda} & 
MIC (HRDA) & Ground Truth \\
\end{tabularx}
} %
\includegraphics[width=\linewidth]{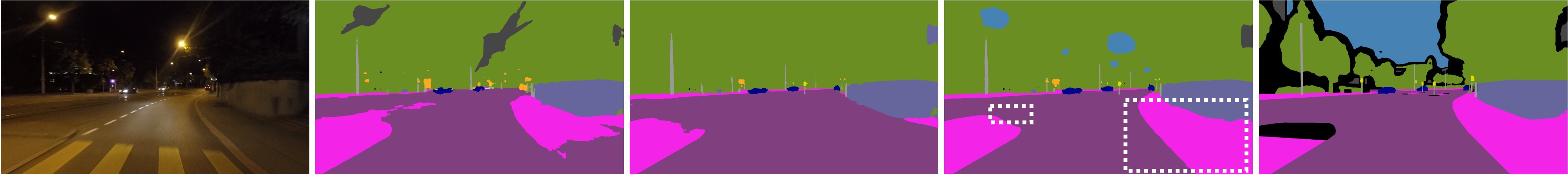}
\includegraphics[width=\linewidth]{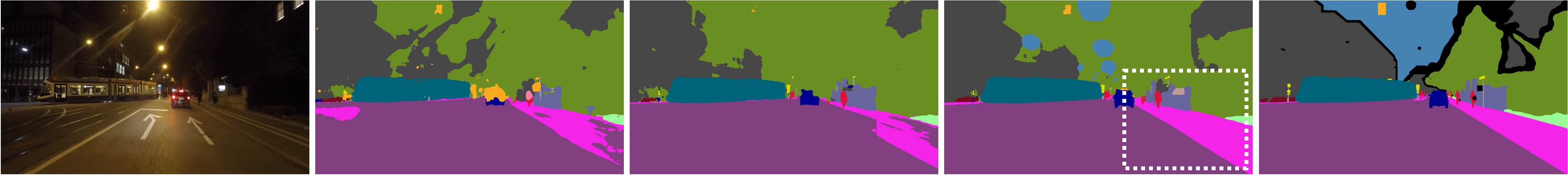}
\includegraphics[width=\linewidth]{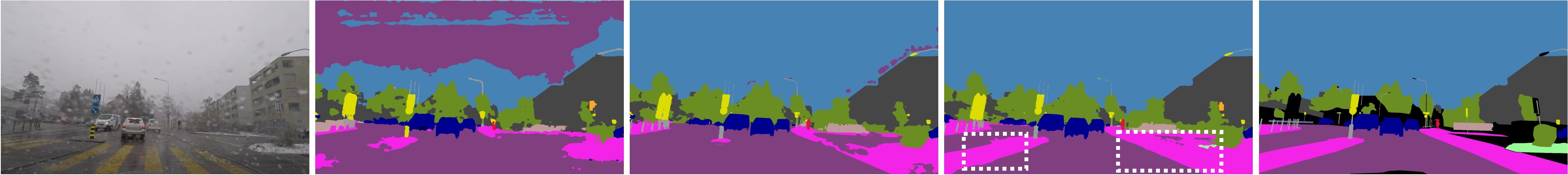}
\scriptsize%
\setlength\tabcolsep{1pt}%
{%
\newcolumntype{P}[1]{>{\centering\arraybackslash}p{#1}}
\begin{tabular}{@{}*{20}{P{0.09\columnwidth}}@{}}
     {\cellcolor[rgb]{0.5,0.25,0.5}}\textcolor{white}{road} 
     &{\cellcolor[rgb]{0.957,0.137,0.91}}sidew. 
     &{\cellcolor[rgb]{0.275,0.275,0.275}}\textcolor{white}{build.} 
     &{\cellcolor[rgb]{0.4,0.4,0.612}}\textcolor{white}{wall} 
     &{\cellcolor[rgb]{0.745,0.6,0.6}}fence 
     &{\cellcolor[rgb]{0.6,0.6,0.6}}pole 
     &{\cellcolor[rgb]{0.98,0.667,0.118}}tr. light
     &{\cellcolor[rgb]{0.863,0.863,0}}tr. sign 
     &{\cellcolor[rgb]{0.42,0.557,0.137}}veget. 
     &{\cellcolor[rgb]{0.596,0.984,0.596}}terrain 
     &{\cellcolor[rgb]{0.275,0.510,0.706}}sky
     &{\cellcolor[rgb]{0.863,0.078,0.235}}\textcolor{white}{person} 
     &{\cellcolor[rgb]{0.988,0.494,0.635}}\textcolor{black}{rider} 
     &{\cellcolor[rgb]{0,0,0.557}}\textcolor{white}{car} 
     &{\cellcolor[rgb]{0,0,0.275}}\textcolor{white}{truck} 
     &{\cellcolor[rgb]{0,0.235,0.392}}\textcolor{white}{bus}
     &{\cellcolor[rgb]{0,0.392,0.471}}\textcolor{white}{train} 
     &{\cellcolor[rgb]{0,0,0.902}}\textcolor{white}{m.bike} 
     & {\cellcolor[rgb]{0.467,0.043,0.125}}\textcolor{white}{bike}
     &{\cellcolor[rgb]{0,0,0}}\textcolor{white}{n/a.}
\end{tabular}
}%

\vspace{-0.15cm}
\caption{Example predictions showing a better segmentation of \emph{sidewalk} by MIC on Cityscapes$\rightarrow$ACDC.}
\label{fig:predictions_acdc_sidewalk}
\end{figure*}

\begin{figure*}
\centering
{\footnotesize
\begin{tabularx}{\linewidth}{*{6}{Y}}
Image & 
DAFormer~\cite{hoyer2021daformer} & 
HRDA~\cite{hoyer2022hrda} & 
MIC (HRDA) & Ground Truth \\
\end{tabularx}
} %
\includegraphics[width=\linewidth]{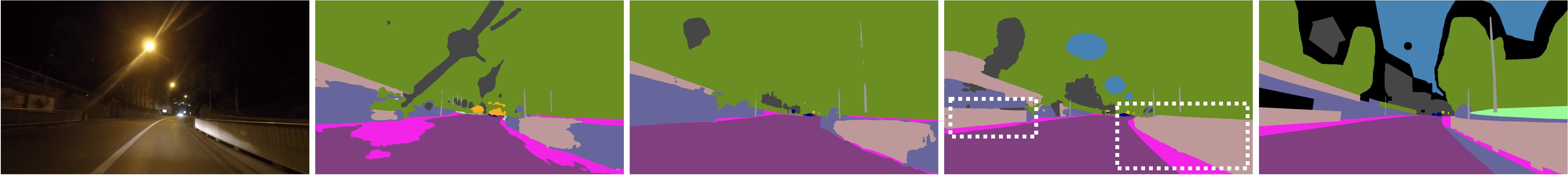}
\includegraphics[width=\linewidth]{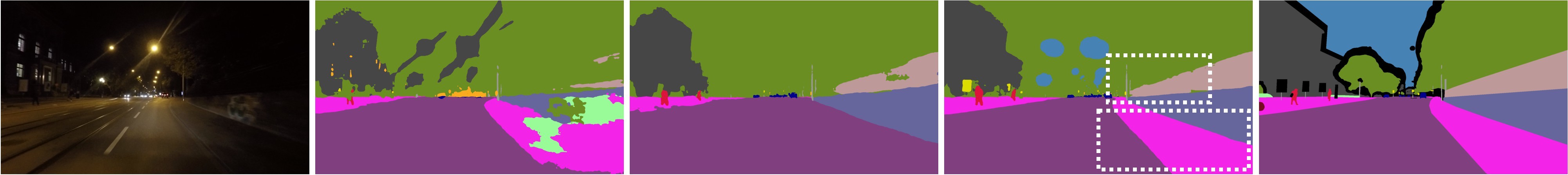}
\includegraphics[width=\linewidth]{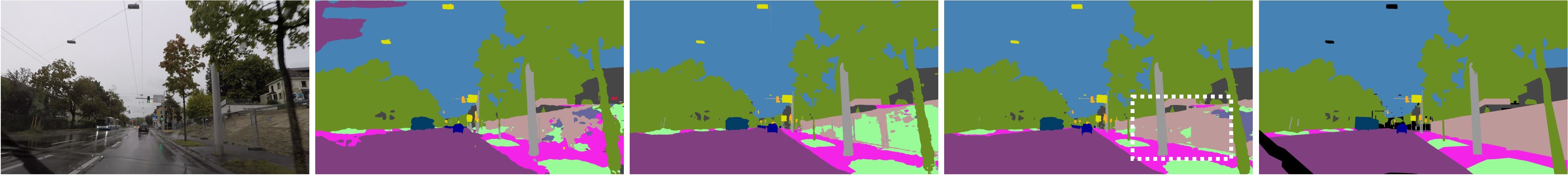}
\scriptsize%
\setlength\tabcolsep{1pt}%
{%
\newcolumntype{P}[1]{>{\centering\arraybackslash}p{#1}}
\begin{tabular}{@{}*{20}{P{0.09\columnwidth}}@{}}
     {\cellcolor[rgb]{0.5,0.25,0.5}}\textcolor{white}{road} 
     &{\cellcolor[rgb]{0.957,0.137,0.91}}sidew. 
     &{\cellcolor[rgb]{0.275,0.275,0.275}}\textcolor{white}{build.} 
     &{\cellcolor[rgb]{0.4,0.4,0.612}}\textcolor{white}{wall} 
     &{\cellcolor[rgb]{0.745,0.6,0.6}}fence 
     &{\cellcolor[rgb]{0.6,0.6,0.6}}pole 
     &{\cellcolor[rgb]{0.98,0.667,0.118}}tr. light
     &{\cellcolor[rgb]{0.863,0.863,0}}tr. sign 
     &{\cellcolor[rgb]{0.42,0.557,0.137}}veget. 
     &{\cellcolor[rgb]{0.596,0.984,0.596}}terrain 
     &{\cellcolor[rgb]{0.275,0.510,0.706}}sky
     &{\cellcolor[rgb]{0.863,0.078,0.235}}\textcolor{white}{person} 
     &{\cellcolor[rgb]{0.988,0.494,0.635}}\textcolor{black}{rider} 
     &{\cellcolor[rgb]{0,0,0.557}}\textcolor{white}{car} 
     &{\cellcolor[rgb]{0,0,0.275}}\textcolor{white}{truck} 
     &{\cellcolor[rgb]{0,0.235,0.392}}\textcolor{white}{bus}
     &{\cellcolor[rgb]{0,0.392,0.471}}\textcolor{white}{train} 
     &{\cellcolor[rgb]{0,0,0.902}}\textcolor{white}{m.bike} 
     & {\cellcolor[rgb]{0.467,0.043,0.125}}\textcolor{white}{bike}
     &{\cellcolor[rgb]{0,0,0}}\textcolor{white}{n/a.}
\end{tabular}
}%

\vspace{-0.15cm}
\caption{Example predictions showing a better segmentation of \emph{fence} by MIC on Cityscapes$\rightarrow$ACDC.}
\label{fig:predictions_acdc_fence}
\end{figure*}

\begin{figure*}
\centering
{\footnotesize
\begin{tabularx}{\linewidth}{*{6}{Y}}
Image & 
DAFormer~\cite{hoyer2021daformer} & 
HRDA~\cite{hoyer2022hrda} & 
MIC (HRDA) & Ground Truth \\
\end{tabularx}
} %
\includegraphics[width=\linewidth]{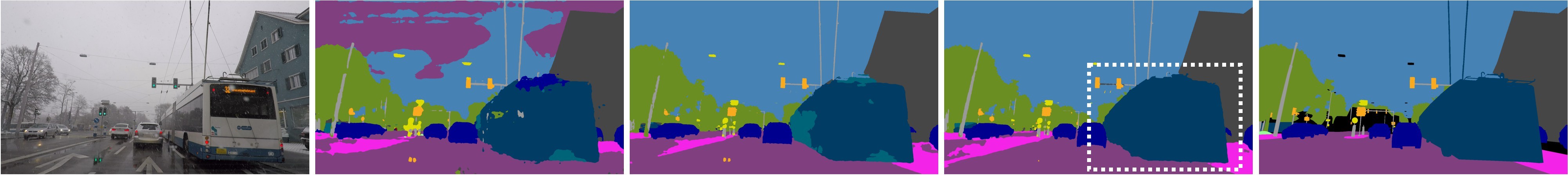}
\includegraphics[width=\linewidth]{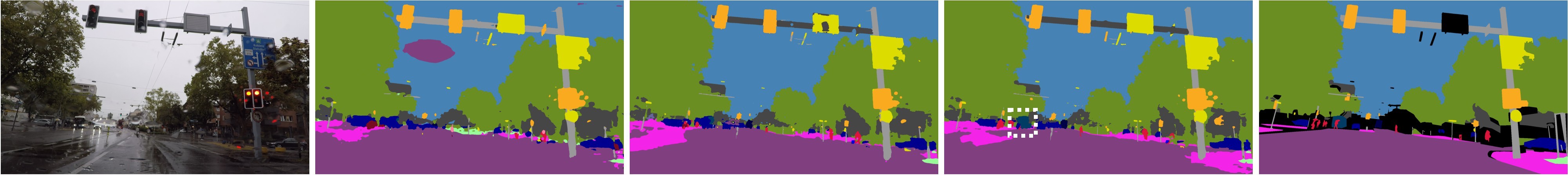}
\includegraphics[width=\linewidth]{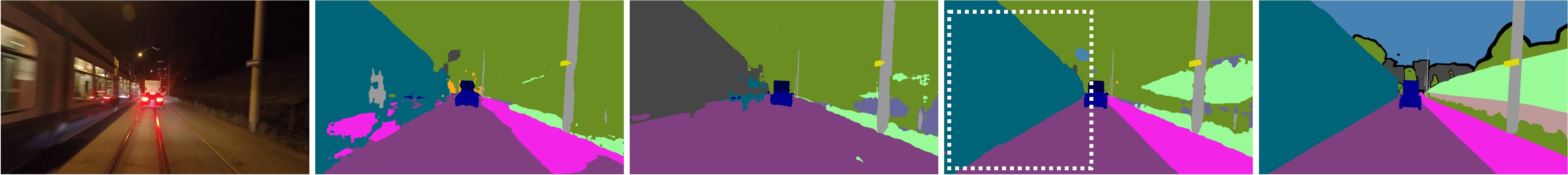}
\scriptsize%
\setlength\tabcolsep{1pt}%
{%
\newcolumntype{P}[1]{>{\centering\arraybackslash}p{#1}}
\begin{tabular}{@{}*{20}{P{0.09\columnwidth}}@{}}
     {\cellcolor[rgb]{0.5,0.25,0.5}}\textcolor{white}{road} 
     &{\cellcolor[rgb]{0.957,0.137,0.91}}sidew. 
     &{\cellcolor[rgb]{0.275,0.275,0.275}}\textcolor{white}{build.} 
     &{\cellcolor[rgb]{0.4,0.4,0.612}}\textcolor{white}{wall} 
     &{\cellcolor[rgb]{0.745,0.6,0.6}}fence 
     &{\cellcolor[rgb]{0.6,0.6,0.6}}pole 
     &{\cellcolor[rgb]{0.98,0.667,0.118}}tr. light
     &{\cellcolor[rgb]{0.863,0.863,0}}tr. sign 
     &{\cellcolor[rgb]{0.42,0.557,0.137}}veget. 
     &{\cellcolor[rgb]{0.596,0.984,0.596}}terrain 
     &{\cellcolor[rgb]{0.275,0.510,0.706}}sky
     &{\cellcolor[rgb]{0.863,0.078,0.235}}\textcolor{white}{person} 
     &{\cellcolor[rgb]{0.988,0.494,0.635}}\textcolor{black}{rider} 
     &{\cellcolor[rgb]{0,0,0.557}}\textcolor{white}{car} 
     &{\cellcolor[rgb]{0,0,0.275}}\textcolor{white}{truck} 
     &{\cellcolor[rgb]{0,0.235,0.392}}\textcolor{white}{bus}
     &{\cellcolor[rgb]{0,0.392,0.471}}\textcolor{white}{train} 
     &{\cellcolor[rgb]{0,0,0.902}}\textcolor{white}{m.bike} 
     & {\cellcolor[rgb]{0.467,0.043,0.125}}\textcolor{white}{bike}
     &{\cellcolor[rgb]{0,0,0}}\textcolor{white}{n/a.}
\end{tabular}
}%

\vspace{-0.15cm}
\caption{Example predictions showing a better segmentation of \emph{bus} and \emph{train} by MIC on Cityscapes$\rightarrow$ACDC.}
\label{fig:predictions_acdc_bus}
\end{figure*}

\begin{figure*}
\centering
{\footnotesize
\begin{tabularx}{\linewidth}{*{6}{Y}}
Image & 
DAFormer~\cite{hoyer2021daformer} & 
HRDA~\cite{hoyer2022hrda} & 
MIC (HRDA) & Ground Truth \\
\end{tabularx}
} %
\includegraphics[width=\linewidth]{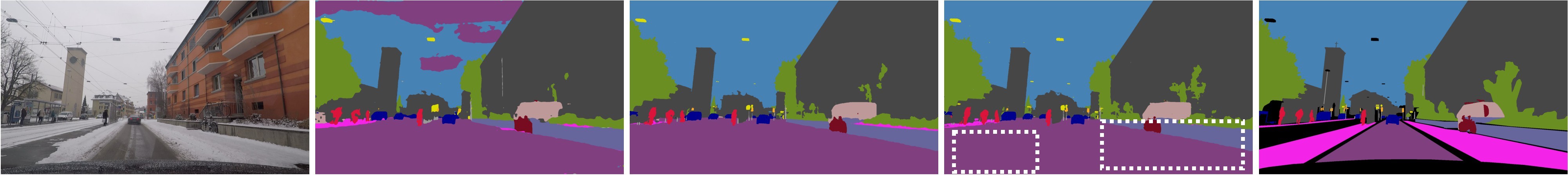}
\includegraphics[width=\linewidth]{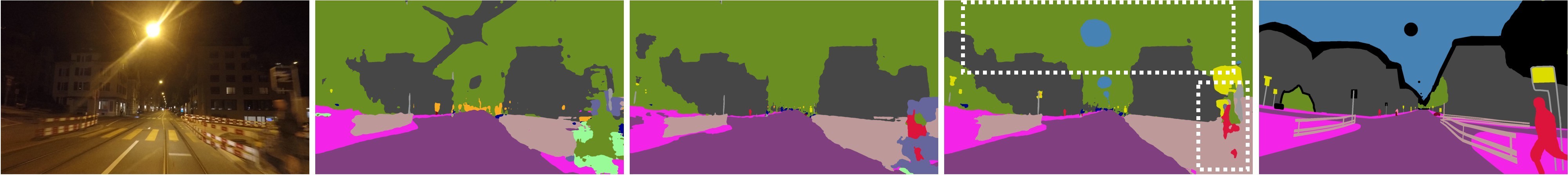}
\scriptsize%
\setlength\tabcolsep{1pt}%
{%
\newcolumntype{P}[1]{>{\centering\arraybackslash}p{#1}}
\begin{tabular}{@{}*{20}{P{0.09\columnwidth}}@{}}
     {\cellcolor[rgb]{0.5,0.25,0.5}}\textcolor{white}{road} 
     &{\cellcolor[rgb]{0.957,0.137,0.91}}sidew. 
     &{\cellcolor[rgb]{0.275,0.275,0.275}}\textcolor{white}{build.} 
     &{\cellcolor[rgb]{0.4,0.4,0.612}}\textcolor{white}{wall} 
     &{\cellcolor[rgb]{0.745,0.6,0.6}}fence 
     &{\cellcolor[rgb]{0.6,0.6,0.6}}pole 
     &{\cellcolor[rgb]{0.98,0.667,0.118}}tr. light
     &{\cellcolor[rgb]{0.863,0.863,0}}tr. sign 
     &{\cellcolor[rgb]{0.42,0.557,0.137}}veget. 
     &{\cellcolor[rgb]{0.596,0.984,0.596}}terrain 
     &{\cellcolor[rgb]{0.275,0.510,0.706}}sky
     &{\cellcolor[rgb]{0.863,0.078,0.235}}\textcolor{white}{person} 
     &{\cellcolor[rgb]{0.988,0.494,0.635}}\textcolor{black}{rider} 
     &{\cellcolor[rgb]{0,0,0.557}}\textcolor{white}{car} 
     &{\cellcolor[rgb]{0,0,0.275}}\textcolor{white}{truck} 
     &{\cellcolor[rgb]{0,0.235,0.392}}\textcolor{white}{bus}
     &{\cellcolor[rgb]{0,0.392,0.471}}\textcolor{white}{train} 
     &{\cellcolor[rgb]{0,0,0.902}}\textcolor{white}{m.bike} 
     & {\cellcolor[rgb]{0.467,0.043,0.125}}\textcolor{white}{bike}
     &{\cellcolor[rgb]{0,0,0}}\textcolor{white}{n/a.}
\end{tabular}
}%

\vspace{-0.15cm}
\caption{Failure cases of MIC on Cityscapes$\rightarrow$ACDC.}
\label{fig:predictions_acdc_failure}
\end{figure*}

\begin{figure*}
\centering
{\footnotesize
\begin{tabularx}{\linewidth}{*{4}{Y}}
SIGMA~\cite{li2022sigma} & 
SADA~\cite{chen2021scale} & 
MIC (SADA) & 
Ground Truth \\
\end{tabularx}
} %
\includegraphics[width=\linewidth]{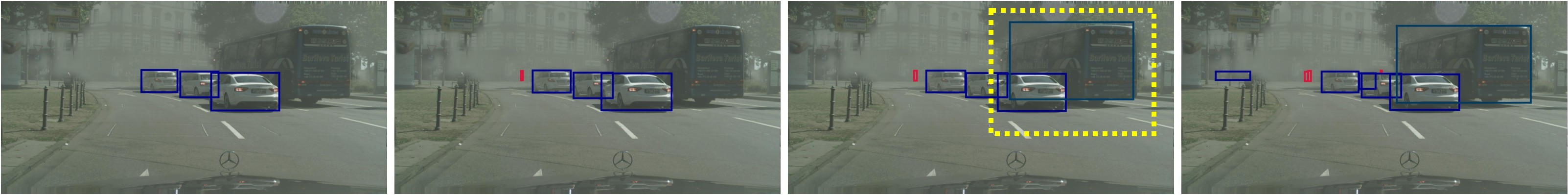}
\includegraphics[width=\linewidth]{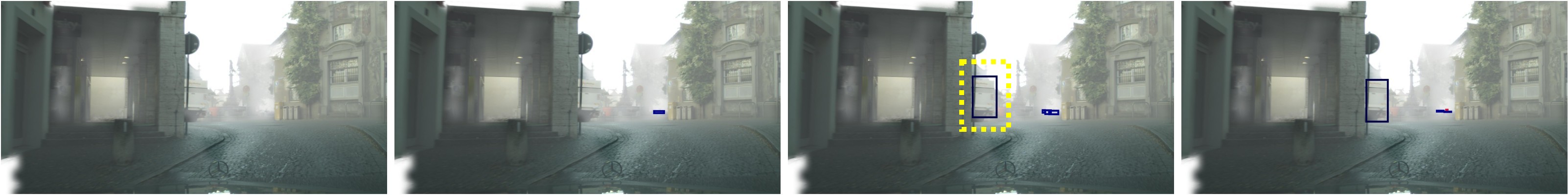}
\includegraphics[width=\linewidth]{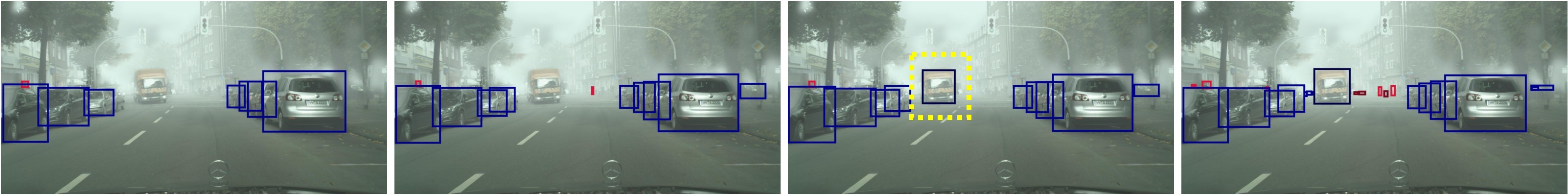}
\scriptsize%
\setlength\tabcolsep{1pt}%
{%
\newcolumntype{P}[1]{>{\centering\arraybackslash}p{#1}}
\begin{tabular}{@{}*{20}{P{0.09\columnwidth}}@{}}
     &{\cellcolor[rgb]{0.863,0.078,0.235}}\textcolor{white}{person} 
     &{\cellcolor[rgb]{0.988,0.494,0.635}}\textcolor{black}{rider} 
     &{\cellcolor[rgb]{0,0,0.557}}\textcolor{white}{car} 
     &{\cellcolor[rgb]{0,0,0.275}}\textcolor{white}{truck} 
     &{\cellcolor[rgb]{0,0.235,0.392}}\textcolor{white}{bus}
     &{\cellcolor[rgb]{0,0.392,0.471}}\textcolor{white}{train} 
     &{\cellcolor[rgb]{0,0,0.902}}\textcolor{white}{m.bike} 
     &{\cellcolor[rgb]{0.467,0.043,0.125}}\textcolor{white}{bike}
\end{tabular}
}%

\vspace{-0.15cm}
\caption{Example predictions showing a better detection of \emph{bus} and \emph{truck} by MIC on Cityscapes$\rightarrow$Foggy Cityscapes.}
\label{fig:predictions_foggy_bus}
\end{figure*}

\begin{figure*}
\centering
{\footnotesize
\begin{tabularx}{\linewidth}{*{4}{Y}}
SIGMA~\cite{li2022sigma} & 
SADA~\cite{chen2021scale} & 
MIC (SADA) & 
Ground Truth \\
\end{tabularx}
} %
\includegraphics[width=\linewidth]{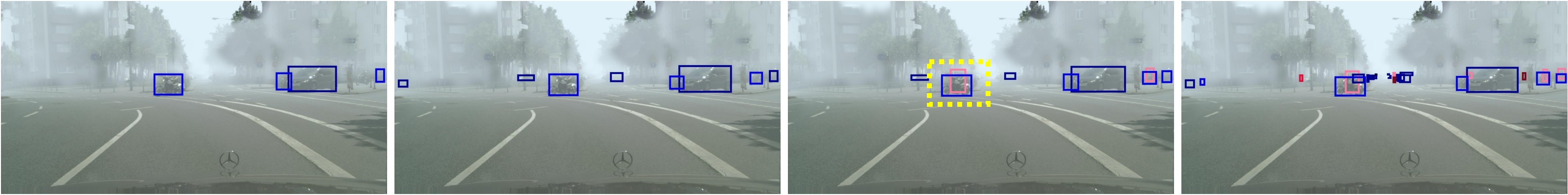}
\includegraphics[width=\linewidth]{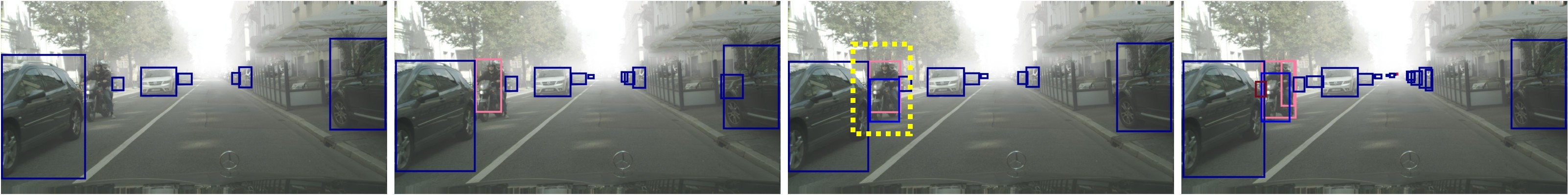}
\includegraphics[width=\linewidth]{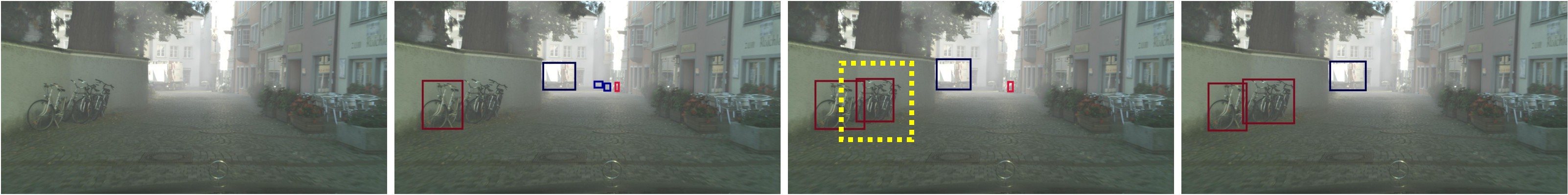}
\scriptsize%
\setlength\tabcolsep{1pt}%
{%
\newcolumntype{P}[1]{>{\centering\arraybackslash}p{#1}}
\begin{tabular}{@{}*{20}{P{0.09\columnwidth}}@{}}
     &{\cellcolor[rgb]{0.863,0.078,0.235}}\textcolor{white}{person} 
     &{\cellcolor[rgb]{0.988,0.494,0.635}}\textcolor{black}{rider} 
     &{\cellcolor[rgb]{0,0,0.557}}\textcolor{white}{car} 
     &{\cellcolor[rgb]{0,0,0.275}}\textcolor{white}{truck} 
     &{\cellcolor[rgb]{0,0.235,0.392}}\textcolor{white}{bus}
     &{\cellcolor[rgb]{0,0.392,0.471}}\textcolor{white}{train} 
     &{\cellcolor[rgb]{0,0,0.902}}\textcolor{white}{m.bike} 
     &{\cellcolor[rgb]{0.467,0.043,0.125}}\textcolor{white}{bike}
\end{tabular}
}%

\vspace{-0.15cm}
\caption{Example predictions showing a better detection of \emph{rider}, \emph{motorcycle}, and \emph{bicycle} by MIC on Cityscapes$\rightarrow$Foggy Cityscapes.}
\label{fig:predictions_foggy_cycle}
\end{figure*}

\begin{figure*}
\centering
{\footnotesize
\begin{tabularx}{\linewidth}{*{4}{Y}}
SIGMA~\cite{li2022sigma} & 
SADA~\cite{chen2021scale} & 
MIC (SADA) & 
Ground Truth \\
\end{tabularx}
} %
\includegraphics[width=\linewidth]{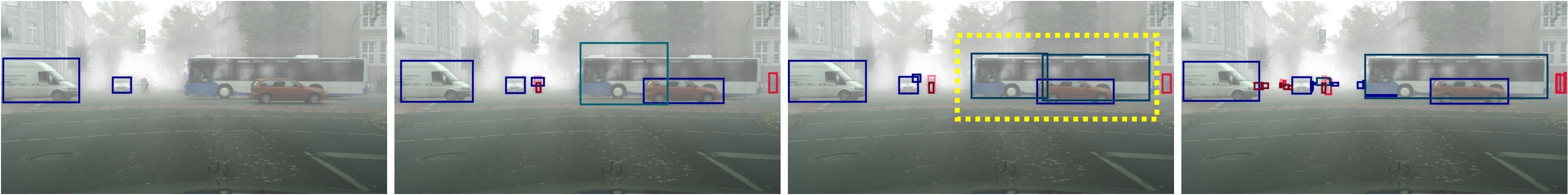}
\includegraphics[width=\linewidth]{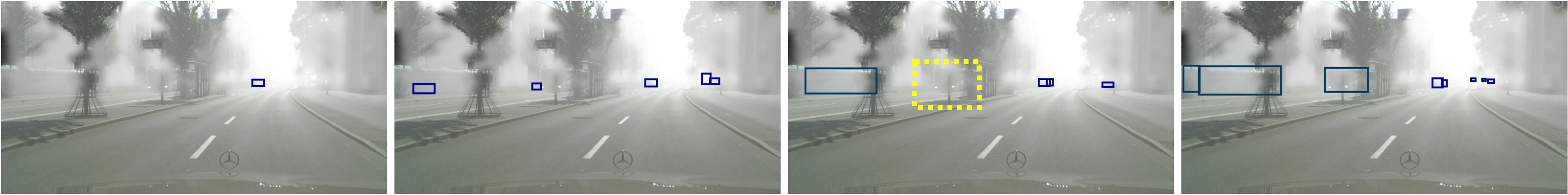}
\includegraphics[width=\linewidth]{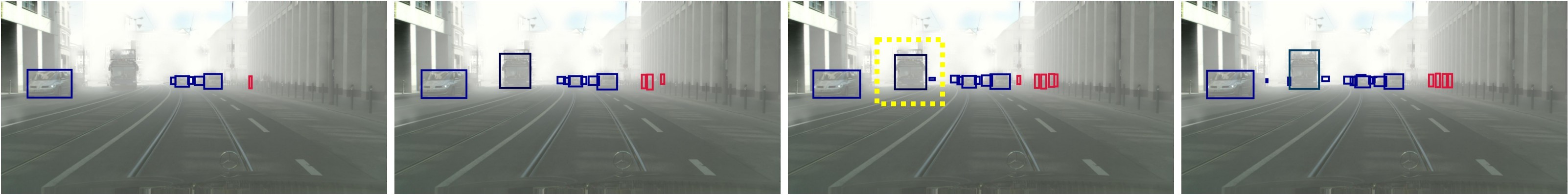}
\scriptsize%
\setlength\tabcolsep{1pt}%
{%
\newcolumntype{P}[1]{>{\centering\arraybackslash}p{#1}}
\begin{tabular}{@{}*{20}{P{0.09\columnwidth}}@{}}
     &{\cellcolor[rgb]{0.863,0.078,0.235}}\textcolor{white}{person} 
     &{\cellcolor[rgb]{0.988,0.494,0.635}}\textcolor{black}{rider} 
     &{\cellcolor[rgb]{0,0,0.557}}\textcolor{white}{car} 
     &{\cellcolor[rgb]{0,0,0.275}}\textcolor{white}{truck} 
     &{\cellcolor[rgb]{0,0.235,0.392}}\textcolor{white}{bus}
     &{\cellcolor[rgb]{0,0.392,0.471}}\textcolor{white}{train} 
     &{\cellcolor[rgb]{0,0,0.902}}\textcolor{white}{m.bike} 
     &{\cellcolor[rgb]{0.467,0.043,0.125}}\textcolor{white}{bike}
\end{tabular}
}%

\vspace{-0.15cm}
\caption{Failure cases of MIC on Cityscapes$\rightarrow$Foggy Cityscapes.}
\label{fig:predictions_foggy_failure}
\end{figure*}

\begin{figure*}
\centering
\includegraphics[width=\linewidth]{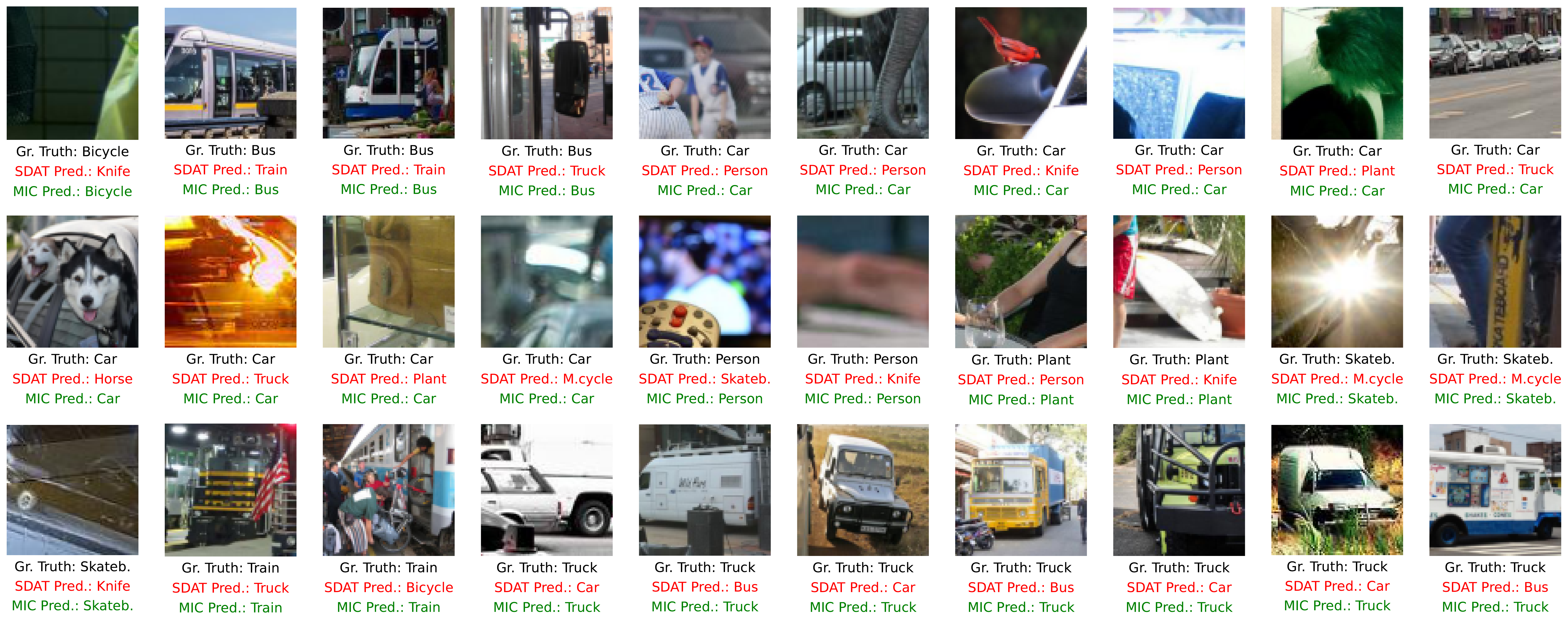}
\vspace{-0.55cm}
\caption{Example predictions showing a better recognition performance of MIC on VisDA.}
\label{fig:predictions_visda}
\end{figure*}

\begin{figure*}
\centering
\includegraphics[width=\linewidth]{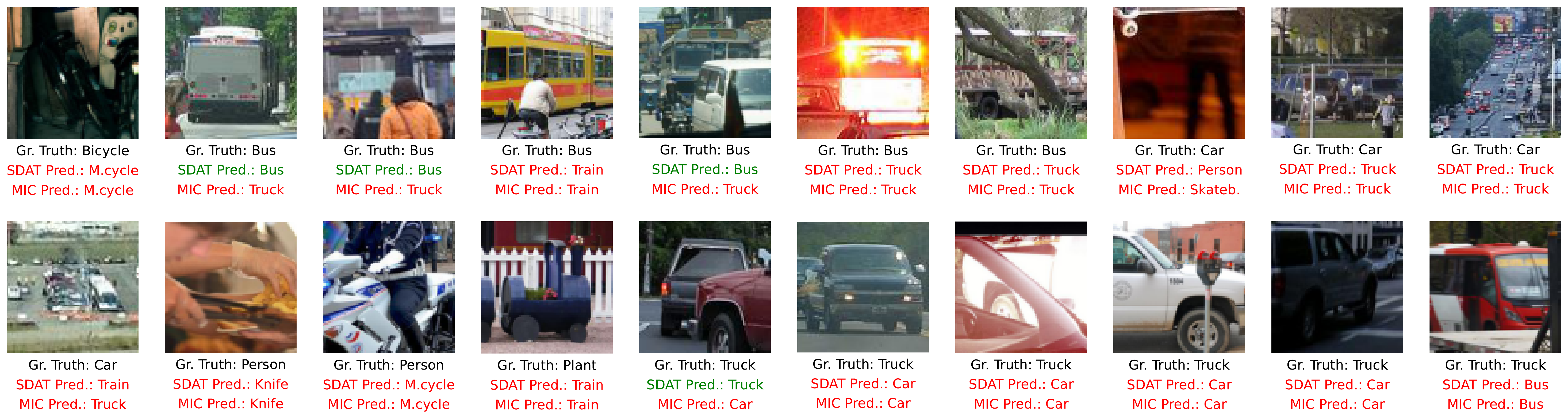}
\vspace{-0.55cm}
\caption{Failure cases of MIC on VisDA.}
\label{fig:predictions_visda_failure}
\end{figure*}

\begin{figure*}
\centering
{\footnotesize
\begin{tabularx}{\linewidth}{*{7}{Y}}
Image & 
CS-Supervised DAFormer~\cite{hoyer2021daformer} & 
CS-Supervised MIC (DAFormer) & Ground Truth \\
\end{tabularx}
} %
\includegraphics[width=\linewidth]{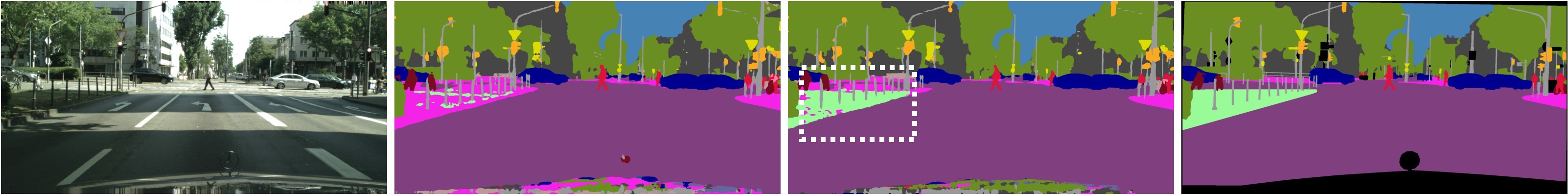}
\includegraphics[width=\linewidth]{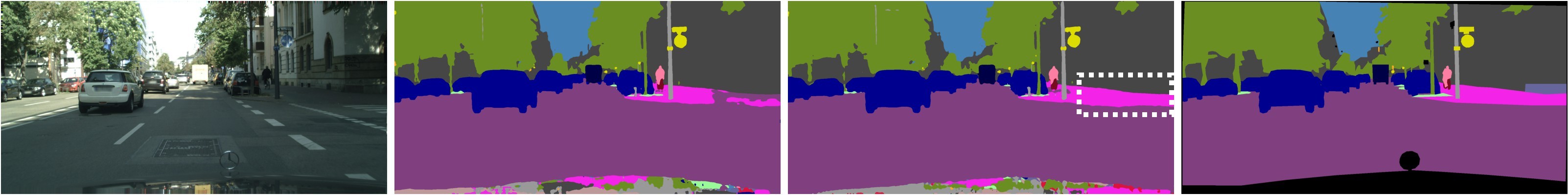}
\includegraphics[width=\linewidth]{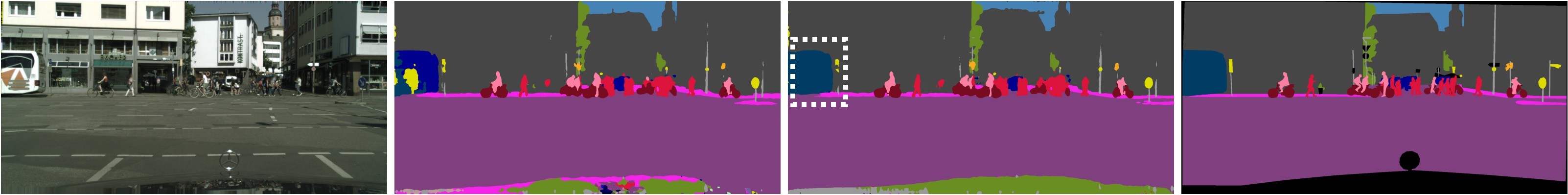}
\includegraphics[width=\linewidth]{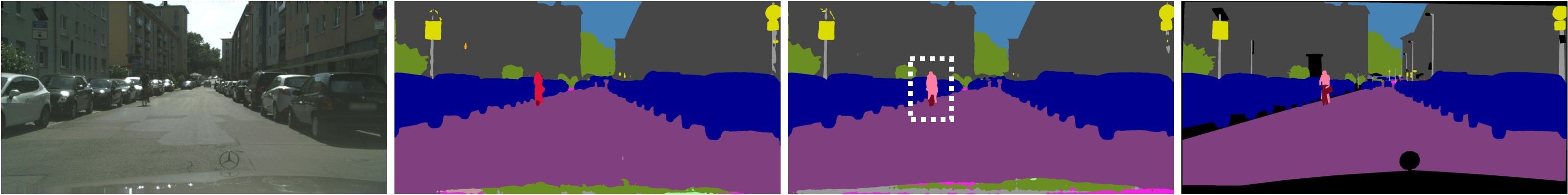}
\scriptsize%
\setlength\tabcolsep{1pt}%
{%
\newcolumntype{P}[1]{>{\centering\arraybackslash}p{#1}}
\begin{tabular}{@{}*{20}{P{0.09\columnwidth}}@{}}
     {\cellcolor[rgb]{0.5,0.25,0.5}}\textcolor{white}{road} 
     &{\cellcolor[rgb]{0.957,0.137,0.91}}sidew. 
     &{\cellcolor[rgb]{0.275,0.275,0.275}}\textcolor{white}{build.} 
     &{\cellcolor[rgb]{0.4,0.4,0.612}}\textcolor{white}{wall} 
     &{\cellcolor[rgb]{0.745,0.6,0.6}}fence 
     &{\cellcolor[rgb]{0.6,0.6,0.6}}pole 
     &{\cellcolor[rgb]{0.98,0.667,0.118}}tr. light
     &{\cellcolor[rgb]{0.863,0.863,0}}tr. sign 
     &{\cellcolor[rgb]{0.42,0.557,0.137}}veget. 
     &{\cellcolor[rgb]{0.596,0.984,0.596}}terrain 
     &{\cellcolor[rgb]{0.275,0.510,0.706}}sky
     &{\cellcolor[rgb]{0.863,0.078,0.235}}\textcolor{white}{person} 
     &{\cellcolor[rgb]{0.988,0.494,0.635}}\textcolor{black}{rider} 
     &{\cellcolor[rgb]{0,0,0.557}}\textcolor{white}{car} 
     &{\cellcolor[rgb]{0,0,0.275}}\textcolor{white}{truck} 
     &{\cellcolor[rgb]{0,0.235,0.392}}\textcolor{white}{bus}
     &{\cellcolor[rgb]{0,0.392,0.471}}\textcolor{white}{train} 
     &{\cellcolor[rgb]{0,0,0.902}}\textcolor{white}{m.bike} 
     & {\cellcolor[rgb]{0.467,0.043,0.125}}\textcolor{white}{bike}
     &{\cellcolor[rgb]{0,0,0}}\textcolor{white}{n/a.}
\end{tabular}
}%

\vspace{-0.15cm}
\caption{Example predictions showing a better segmentation of difficult classes such as \emph{terrain}, \emph{sidewalk}, \emph{bus}, and \emph{rider} by MIC in a \textbf{supervised} training setup on Cityscapes.}
\label{fig:predictions_supervised}
\end{figure*}

Supplementing the example predictions in the main paper, we show further representative examples of the strength and weaknesses of MIC in comparison with strong state-of-the-art methods.

\paragraph{Synthetic-to-Real Segmentation:}
On GTA$\to$CS semantic segmentation, MIC(HRDA) achieves considerable performance improvements for the classes \emph{sidewalk}, \emph{fence}, \emph{bus}, and \emph{rider} (see Tab.~\ref{tab:extended_sota_segmentation}). This is also reflected in the example predictions in Fig.~\ref{fig:predictions_gta_sidewalk}-\ref{fig:predictions_gta_rider}. 
In these examples, it can be observed that previous methods often recognize only parts of ambiguous regions while other parts of the same region are misclassified. 
As MIC was trained to utilize context relations, it has learned to reason more holistically about context relations in the images. Therefore, MIC can probably utilize the correctly recognized object parts to resolve the semantics of ambiguous image regions.
More specifically, for \emph{sidewalk} (Fig.~\ref{fig:predictions_gta_sidewalk}), MIC is able to segment \emph{sidewalk} more completely and even recognizes segments that previous methods failed to identify. For \emph{fence} (Fig.~\ref{fig:predictions_gta_fence}), MIC reduces the segmentation of objects behind the fence instead of the fence. For \emph{bus} (Fig.~\ref{fig:predictions_gta_bus}), MIC better segments ambiguous textures inside the bus and better recognizes partly-occluded busses. For \emph{rider} (Fig.~\ref{fig:predictions_gta_rider}), MIC better segments the upper body and head of close riders and is able to recognize distant riders, probably by utilizing the bicycles as a context clue.

However, there are also some difficult examples, where UDA methods including MIC fail to correctly segment the image (Fig.~\ref{fig:predictions_gta_failure}). For example, MIC still struggles to differentiate vehicles with rare appearances, sidewalk that merges with the road, sidewalk under parking cars, and pedestrians standing close to bicycles.

\paragraph{Clear-to-Adverse-Weather Segmentation:}
On CS$\to$\allowbreak ACDC semantic segmentation, the same observations as for GTA$\to$CS apply for the classes \emph{sidewalk} (Fig.~\ref{fig:predictions_acdc_sidewalk}), \emph{fence} (Fig.~\ref{fig:predictions_acdc_fence}), and \emph{bus/train} (Fig.~\ref{fig:predictions_acdc_bus}). However, there are some distinct failure cases. In particular, UDA methods including MIC fail to segment snow-covered \emph{sidewalk}, distinguish \emph{sky/vegetation/building} in dark image ares, and struggle with motion blur of dynamic objects.

\paragraph{Clear-to-Foggy-Weather Detection:}
On CS$\to$Foggy CS object detection, MIC(SADA) is able to detect objects that previous methods failed to recognize. For example, MIC better detects the classes \emph{bus} and \emph{truck} (Fig.~\ref{fig:predictions_foggy_bus}) as well as \emph{rider}, \emph{motorcycle}, and \emph{bicycle} (Fig.~\ref{fig:predictions_foggy_cycle}). Typical failure cases (Fig.~\ref{fig:predictions_foggy_failure}) include multiple detections for a single object, missed detections, and the confusion of semantically similar objects.

\paragraph{Synthetic-to-Real Classification:}
For VisDA-2017 image classification UDA, we provide a random selection of examples, where MIC(SDAT) performs better than SDAT in Fig.~\ref{fig:predictions_visda}. It can be seen that MIC can better distinguish semantically similar classes such as \emph{train} vs. \emph{bus}, \emph{bus} vs \emph{truck}, and \emph{truck} vs \emph{car}. Further, we show a random selection of failure cases of MIC(SDAT) in Fig.~\ref{fig:predictions_visda_failure}. MIC mostly confuses semantically similar vehicle classes, especially if instances are at the decision boundary between two classes or different classes are present in an image.

\paragraph{Supervised Segmentation:}
Fig.~\ref{fig:predictions_supervised} compares DAFormer and MIC(DAFormer) when trained in a supervised fashion on Cityscapes. It shows improvements for regions that are difficult to identify such as instances of \emph{terrain}, \emph{sidewalk}, \emph{bus}, and \emph{rider}. Generally, the supervised DAFormer without MIC already performs very well, so that the potential for improvement is smaller, which is also reflected in the quantitative results in the main paper.

\section{Potential Limitations}
\label{sec:supp_limitations}

Even though UDA methods achieve evolvingly higher performances for synthetic-to-real and clear-to-adverse weather adaptation, the current methods are still not reliable enough to be safely deployed in real-world autonomous driving as can be seen in the failure cases in Fig.~\ref{fig:predictions_gta_failure}, \ref{fig:predictions_acdc_failure}, and \ref{fig:predictions_foggy_failure}. For these cases, it is still necessary to collect annotations on the target domain to achieve safe operation. We hope that this gap to supervised learning can be gradually narrowed in the future, but we assume that, for some corner cases, a few annotations might still be necessary to reliably guide the adaptation.

As MIC is specifically exploiting context relations for domain adaptation, it is based on two assumptions. First, MIC assumes that context information is a relevant factor for recognition. For classes, where context is less important, such as \emph{building} or \emph{vegetation} for synthetic-to-real adaptation, MIC has a limited potential for improvement. And second, MIC assumes that the relevant context relations are captured by the training data. If objects appear out-of-context during inference, MIC might be more susceptible to these corner cases. 
In the experimental analysis, it is shown that these assumptions mostly hold on a wide range of practically-relevant UDA benchmarks and MIC outperforms previous methods by a significant margin.

\end{document}